\documentclass[twoside]{article}
\usepackage{aistats2018}
\usepackage{color}
\usepackage{dblfloatfix}
\usepackage{amsmath}
\usepackage{amssymb}
\usepackage{graphicx}

\newcommand{\red}[1]{\textcolor{red}{#1}}
\newcommand{\blue}[1]{\textcolor{blue}{#1}}

\newcommand{\sA}{\mathcal{A}}

\begin{document}



\twocolumn[

\aistatstitle{Time Series Using Exponential Smoothing Cells}

\aistatsauthor{ Avner Abrami \And Aleksandr Aravkin \And  Younghun Kim}

\aistatsaddress{IBM TJ Watson Research Center \And  Applied Mathematics \\University of Washington \And Utopus Insights } ]

\begin{abstract}
Time series analysis is used to understand and predict dynamic processes, including evolving demands in business, weather, 
markets, and biological rhythms. Exponential smoothing is used in all these domains to obtain simple interpretable models of time series and to forecast future values. Despite its popularity, exponential smoothing fails dramatically in the presence 
of outliers, large amounts of noise, or when the underlying time series changes. 

We propose a flexible model for time series analysis, using exponential smoothing cells for overlapping time windows. 
The approach can detect and remove outliers, denoise data, fill in missing observations, and provide meaningful forecasts in challenging situations. 
In contrast to classic exponential smoothing, which solves a {\it nonconvex} optimization problem over the smoothing parameters and initial state, the proposed approach requires solving a single structured {\it convex} optimization problem. Recent developments in efficient convex optimization of large-scale dynamic models make the approach tractable. We illustrate new capabilities using
synthetic examples, and then use the approach to analyze and forecast noisy real-world time series.  Code for the approach and experiments is publicly available.
\end{abstract}

\section{Introduction}

Exponential smoothing (ES) methods model current and future time series observations as 
a weighted combinations of past observations, with more weight given to recent data. 
The word `exponential' reflects the exponential decay of weights for older observations. 
ES methods have been around since the 1950s, 
and are still very popular forecasting methods used in business and industry, 
including supply chain forecasting~\cite{chen2000impact}, 
stock market analysis~\cite{taylor2004volatility,pindyck1983risk,brown1961fundamental},
weather prediction~\cite{taylor2002neural,soman2010review}, and electricity demand forecasting~\cite{taylor2003short,moghram1989analysis}. 

In contrast to many techniques in machine learning, ES provides 
simple and interpretable models and forecasting capability 
by assuming a fixed structure for the evolution of the time series.
%for example, level (average), trend (change over time) and seasonality 
%(periodic components with known period) that change in a uniform way  
%over a period of interest.   
For example, a simple (level only) model is
\begin{equation}
\label{eq:simplest}
\hat y_{t+1} = \hat y_t +\alpha(y_t - \hat y_t) = (1-\alpha) \hat y_t + \alpha y_t,
\end{equation}
where $y_t \in \mathbb{R}$ is an observation at time $t$, and $\hat y_t$ is the estimate 
of $y_t$ at time $t$ given $(y_1, \dots, y_{t-1})$. 
The forecast at $t+1$ is adjusted by a fraction $\alpha \in (0,1)$ of the error at time $t$; larger $\alpha$
means greater adjustment.  
%When $\alpha$ is close to $1$, the new forecast is significantly adjusted for the error in the previous forecast; 
%when $\alpha$ is close to $0$, the new forecast is barely adjusted. 
Iterating~\eqref{eq:simplest}, we have 
\begin{equation}
\label{eq:recurr}
\hat y_{t+1} = \alpha\sum_{i = 0}^{t-1} (1-\alpha)^{i} y_{t-i} + (1-\alpha)^t\hat y_1,
\end{equation}
illustrating the exponential decay. 
%When the error scales with the level, 
%a {\it multiplicative} analogue can be used instead: 
%\begin{equation}
%\label{eq:simplestMult}
%\hat y_{t+1} = \hat y_t \left(\frac{y_t}{\hat y_t}\right)^\alpha.
%\end{equation}

\begin{figure*}
%\center
\hspace{-.5cm}
\begin{tabular}{c}\\
\includegraphics[scale=0.6]{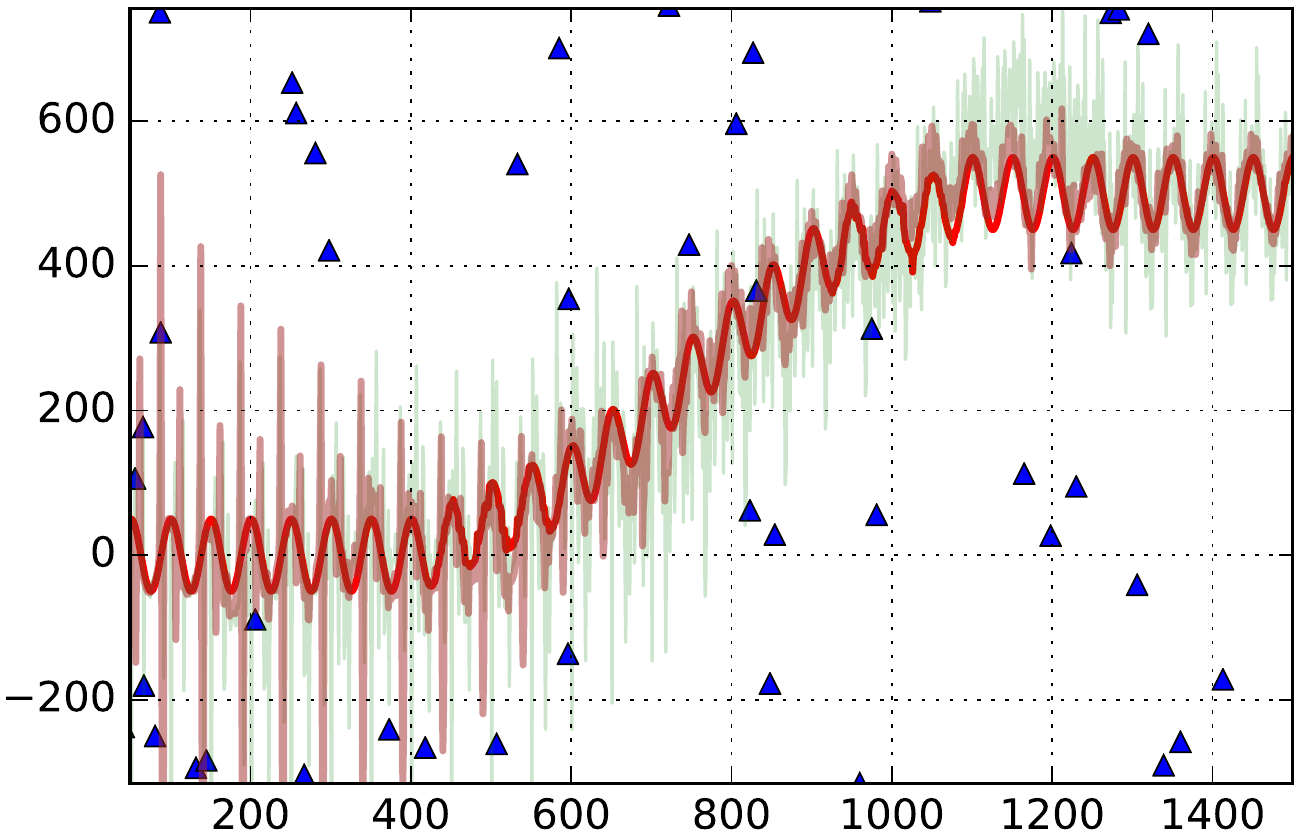} \\
(a) {\bf Noisy series fit by HW (green), Robust }\\
{\bf HW (red transparent), and ES cell (red solid).}
\end{tabular}
\begin{tabular}{cc}\\ 
\includegraphics[scale=0.3]{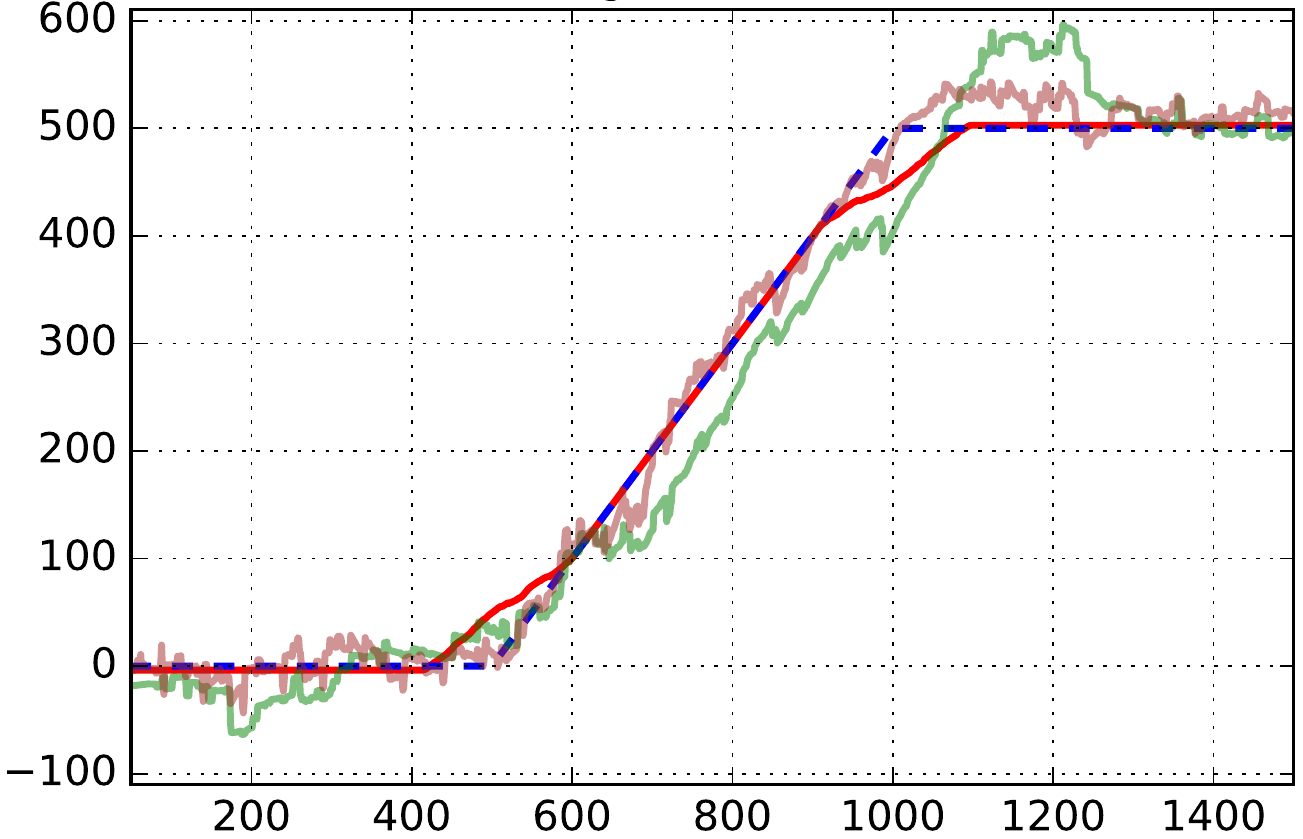}
& 
\includegraphics[scale=0.3]{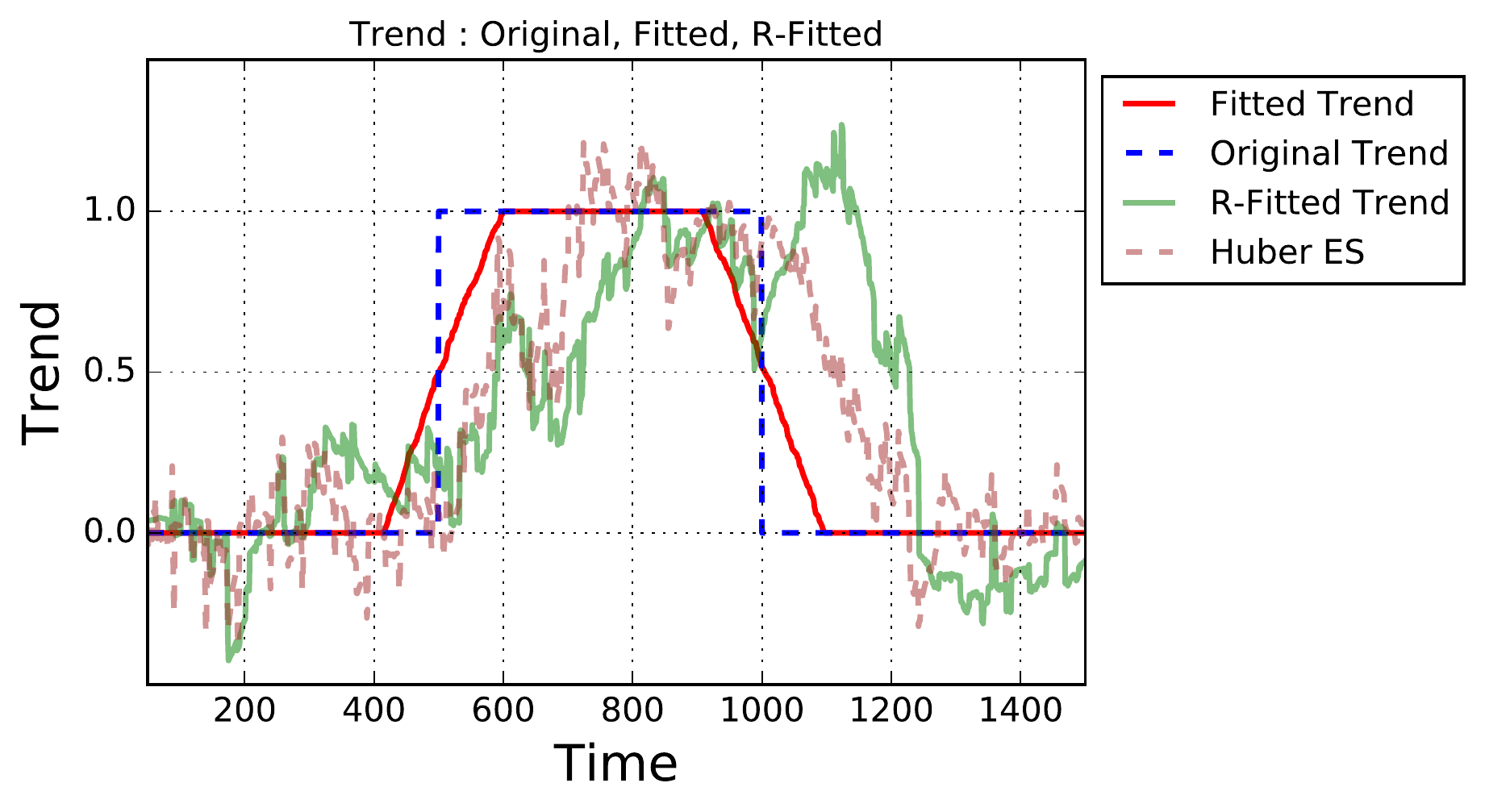}
\\ 
(b) {\bf Level } 
\hspace{-.2in}
& 
(c) {\bf Trend}\\
\includegraphics[scale=0.3]{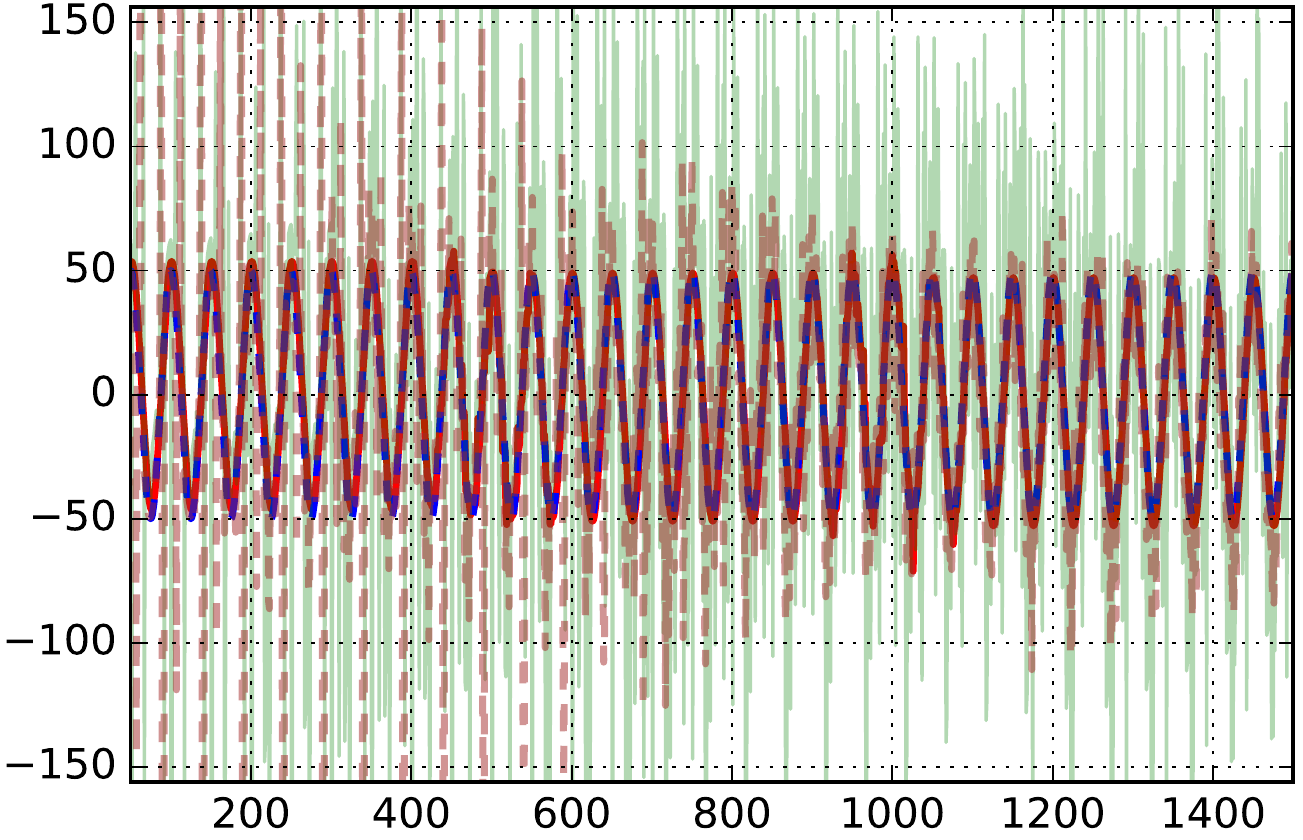}
&\includegraphics[scale=0.3]{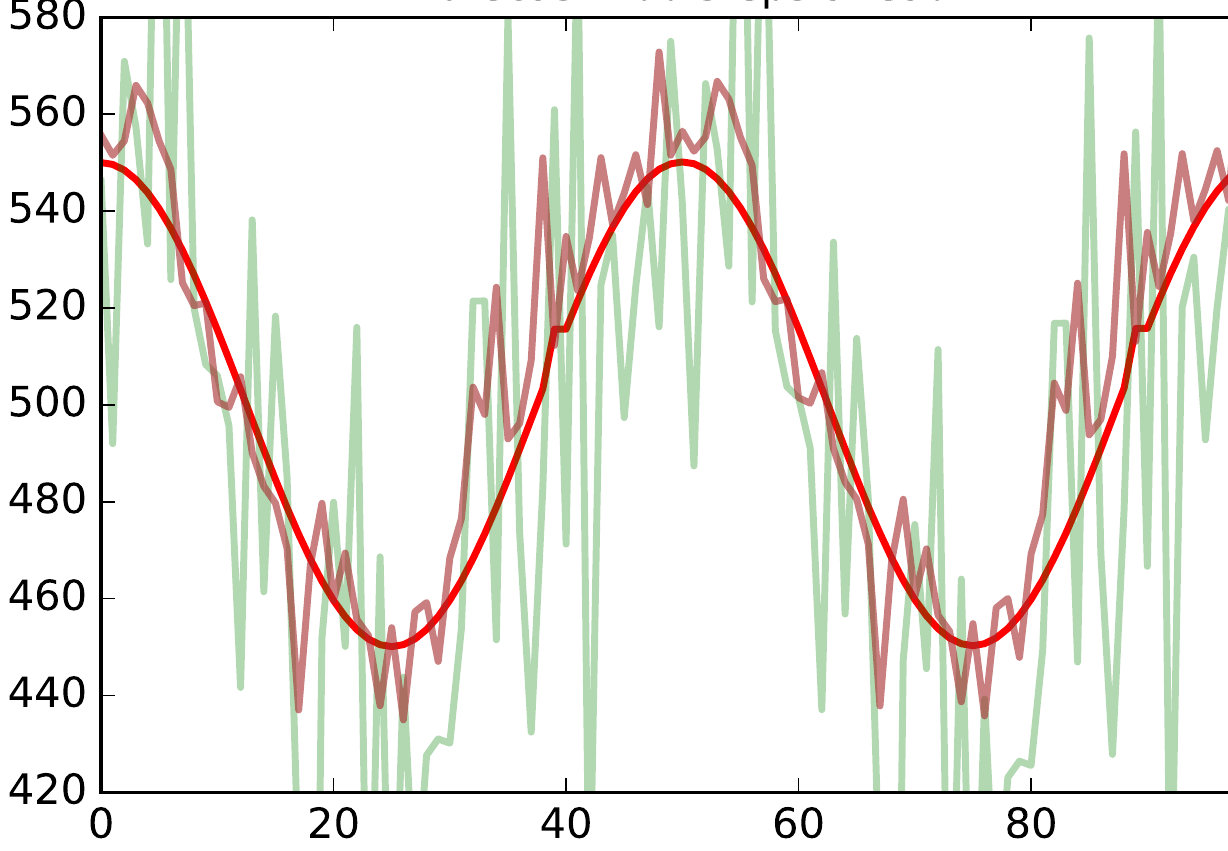}
\\
(d) {\bf Seasonality}  & (e) {\bf Forecast (100 steps)}.
%(e) {\bf Outliers detected by ES.}
\end{tabular}
\caption{\label{fig:noisy} Limitations of classic ES smoothing (e.g. Holt-Winters and robust variants) are eliminated by the new ES-Cells approach. In all panels, standard Holt-Winters (HW) estimates are shown in green. 
Robust Holt-Winters (RHW)~\cite{gelper2010robust} results, obtained by pre-filtering $y_t$ using a robust loss, are shown 
in transparent red. ES cell results are shown in solid red. 
Panel (a) displays the full time series (outliers plotted as triangles). HW results are noisy; RHW improves after a 
 `burn-in' period; ES cell recovers the underlying time series. 
Panels (b), (c), and (d) compare truth (blue) to estimates of level, trend, and seasonality components. 
Panel (e) shows mean-only forecast for 100 steps ahead.}  
%Right: overall approach, with multiple ES-Cells linked by a dynamic process model.} %(They are given in the text describing the experiment). }
\end{figure*}

The $\alpha$ in~\eqref{eq:simplest} %or~\eqref{eq:simplestMult} 
is fit using available data and used across the entire period of interest. 
More generally, a time series model also includes trend (long-term direction) and periodic seasonality components,
with additional smoothing terms ($\beta, \gamma)$ for these components.  Classic ES methods use observed 
data to fit smoothing components, error variance, and initial state, and then use the quantities to provide point forecasts and quantify uncertainty. 
Every additive ES can be formulated using the compact single source of error (SSOE) representation~\cite{hyndman2002state,hyndman2008forecasting}:
\begin{equation}
\label{eq:ESmodel}
\begin{aligned}
y_{t} &=  w^{T} x_{t-1} + \epsilon_{t}\\ 
x_{t} &= A x_{t-1} + g \epsilon_t 
\end{aligned}
\end{equation}
where $w$ is a linear measurement model, $A$ is a linear transition function and $g$ is the vector of smoothing parameters. In~\eqref{eq:simplest}, we have $w = 1$, $x_t = y_t$, and $A = I$, and $g = \alpha$. 
More generally, $x_t$ tracks the deterministic components of the time series (level, trend, seasonality) while $g$ adjusts for stochastic disturbances.

\section*{Flexibility of ES models}
To show how ES models are constructed and transformed into~\eqref{eq:ESmodel}, 
we compare the simple linear, Holt's linear, and Holt-Winters models.
The simple linear model from~\eqref{eq:simplest}
%re-written in~\eqref{eq:HoltBasicModel}, 
tracks the level $l_t\in \mathbb{R}$ using a zero-order polynomial approximation:
\begin{equation}
\label{eq:HoltBasicModel}
\begin{array}{l}
y_t = l_{t-1} +\epsilon_t \\
l_t = \alpha y_t  + (1-\alpha) l_{t-1}.
\end{array}\quad 
\end{equation}
%At any point along the data path, 
%values at nearby points are approximated by a constant, referred to as a {\it local level}. 
%As the level height changes over time, many local levels are used, see left panel of Figure 2. 
Holt's Linear Model %, shown in~\eqref{eq:HoltLinearModel}, 
uses a first order (tangent) line approximation, 
and tracks both level $l_t$ and trend 
$b_t \in \mathbb{R}$: 
\begin{equation}
\label{eq:HoltLinearModel}
\begin{array}{l}
y_t = l_{t-1} + \blue{b_{t-1}}  + \epsilon_t \\
l_t = \alpha y_t  + (1-\alpha) (l_{t-1} + \blue{b_{t-1}})\\
\blue{b_t = \beta (l_t - l_{t-1}) + (1- \beta) b_{t-1} }.
\end{array}
\end{equation}
%, and the right panel of Figure 2. 
Finally, the Holt-Winters model %, shown in~\eqref{eq:HoltWintersModel}, 
adds a seasonality component $s_t \in \mathbb{R}^p$ with known periodicity $p$, 
giving the augmented state $x_t = (l_t, b_t, s_{t}, s_{t-1}, \dots, s_{t-p-1})$, see~\eqref{eq:HoltWintersModel}.

\begin{equation}
\label{eq:HoltWintersModel}
\begin{array}{l}
y_t = l_{t-1} + \blue{b_{t-1}} + \red{s_{t-p-1}} + \epsilon_t 
\medskip
\\
l_t = \alpha (y_t - \red{s_{t-p-1}}) + (1-\alpha) (l_{t-1} + \blue{b_{t-1}})\\
\blue{b_t = \beta (l_t - l_{t-1}) + (1- \beta) b_{t-1} }\\
\red{s_t = \gamma (y_t - l_{t-1} - b_{t-1}) +(1-\gamma) s_{t-p-1}}.
\end{array} 
\end{equation}

To write the Holt-Winters model in form of~\eqref{eq:ESmodel}, take  
\begin{equation}
\label{eq:SSHoltWinters}
\small
%x_t &= \begin{bmatrix} l_t \\ b_t \\ \epsilon_t \\ s_{t-1} \\ \vdots \\ s_{t-p+1} \\ s_{t-p} \end{bmatrix}, \;\;
w = \begin{bmatrix} 1\\ 1 \\ 0 \\ \vdots \\ 0 \\ 1\end{bmatrix},  
g = \begin{bmatrix} \alpha \\ \beta\\ \gamma \\ 0\\ \vdots \\0 \end{bmatrix}, 
A = \begin{bmatrix} 1 & 1 & 0 & 0 & \dots & 0 &0\\ 
0 & 1 & 0 & 0 & \dots &  0 &0 \\
0 & 0 & 0 & 0 & \dots & 0 &1 \\
0 & 0 & 0 & 1 & \dots & 0 & 0 \\
\vdots & \vdots &  \vdots & \vdots & \ddots& &\vdots\\
0 & 0 & 0 & 0 & & \ddots & 0\\
0 & 0 & 0 &0 & 0 & \dots & 1
\end{bmatrix}.
\end{equation}

\begin{figure*}
\center
\begin{tabular}{cc}\\ 
\includegraphics[scale=0.29]{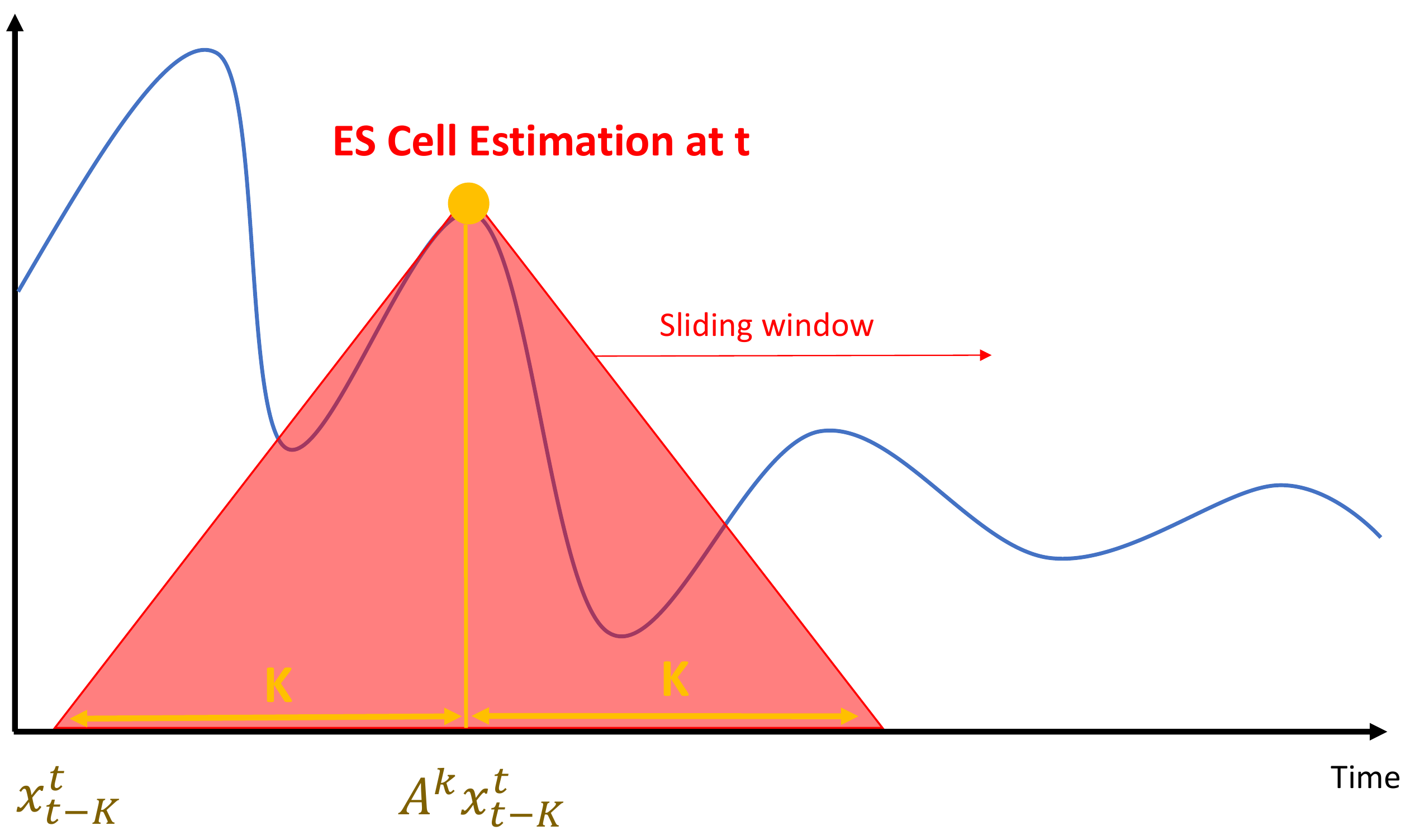}
&\includegraphics[scale=0.29]{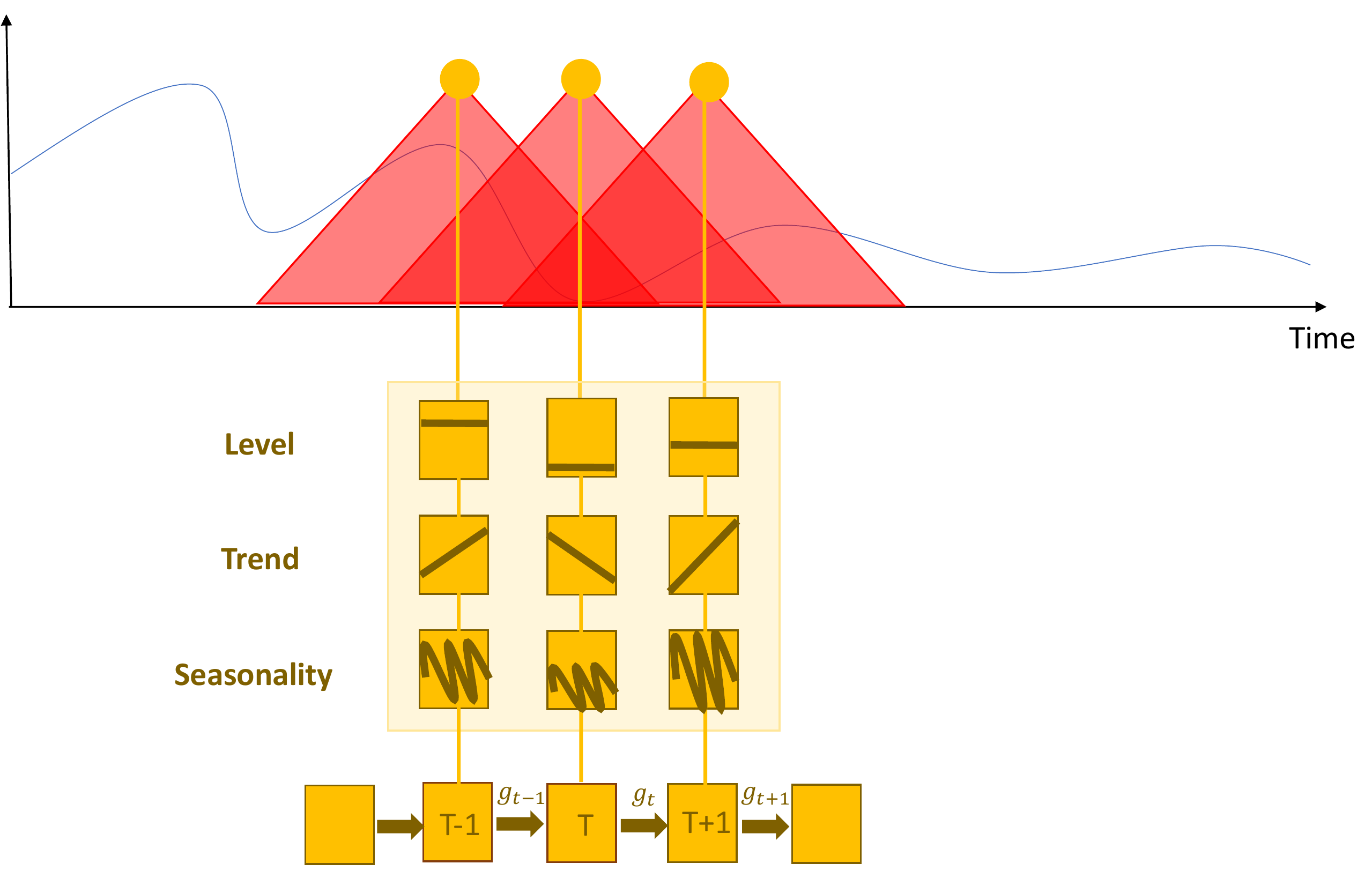}\\
(a) {\bf Local estimation by a single ES cell} 
&(b) {\bf Overall approach: ES-Cells linked by a dynamic model.}
\end{tabular}
\caption{\label{fig:approach} Time series using ES-Cells. Left: a single ES cell, estimating parameters over a particular window of size $2K$. 
Right: overall approach, with multiple ES-Cells linked by a dynamic process model.} %(They are given in the text describing the experiment). }
\end{figure*}

\section*{Limitations of classic ES}
SSOE models are attractive in their simplicity, since they use a single parameter vector $g$ 
to model perturbations. However, the same $g$ must applied at every time point; this limits model 
flexibility and ensures estimates of $g$ are strongly affected by artifacts in the data, including noise and outliers. 
Consider a toy example~\eqref{eq:ESmodel} with two measurements; $g$ and $x_0$ are found by solving 
\begin{equation}
\label{eq:simpleObj}
 \min_{x_{0},g} (y_{1} - w^Tx_{0})^2 + (y_{2} - w^{T}Ax_{0} + w^{T}g(y_{1} - w^{T}x_{0}))^{2}.
 \end{equation}
If $y_1$ is  an outlier, it affects both $x_0$ and $g$ in the fit. Robust statistics are powerless
for SSOE models: ignoring $y_1$ necessarily leaves a large $\epsilon_1 = y_1 - w ^Tx_0$, 
which affects $\epsilon_2$.  The propagation is recursive:  
$$ \epsilon_{3} = y_{3} - w^{T}A^{2}x_{0} + w^{T}A g \epsilon_{1} + w^{T} g \epsilon_{2}. $$
 Each $\epsilon_t$ appears in all $\epsilon_{t+k}$, $k\geq 1$, so an outlier at any point has to affect $x_0$ 
 and $g$, which control the entire time series.  This phenomenon does not occur in other time series 
 formulations, including AR models, where robust methods have been developed,  see e.g.~\cite{maronna2006robust}.
 % the estimates of $x_0$ and $g$. 

Earlier work in robust Holt-Winters (RHW) uses M-estimators to filter the observations $y_t$~\cite{cipra1992robust,gelper2010robust}, and then applies standard HW. 
In contrast, ES-Cells formulates a single problem to analyze the entire time series,
simultaneously denoising, decomposing, and imputing.  

The classic approach~\eqref{eq:simpleObj} is nonconvex, and has weak guarantees: stationarity conditions (such as $\nabla f (x) = 0$) do not imply global optimality, and solutions found by iterative methods depend on the initial point. As the level of noise and outliers increases, the ability of black-box optimizers to get reasonable $x_0$ and $(\alpha, \beta,\gamma)$ breaks down. The ES-Cells approach uses a {\it strongly convex} formulation; it has a unique global minimum and no other stationary points.

We created a synthetic time series with trend and level shifts, as well as heteroscedastic noise and outliers in the observations. 
Figure~\ref{fig:noisy} compares the performance of the HW model\footnote{Implemented in the standard Holt-Winters R module}~\cite{holt2004forecasting} and RHW~\cite{gelper2010robust}\footnote{We implemented the approach. In\cite[Eqs.~13,14]{gelper2010robust}, we take $\sigma_0 = 0.05$, 
and $\lambda_\sigma = 0.01$. Automated methods for 
obtaining smoothing parameters and $x_0$ failed in the presence of noise; so for 
the HW model, we use hand-tuned parameters $\alpha = 0.05$,  
$\beta = 0.01$,  $\gamma = 0.15$, with $x_0$ the first 50 elements of the noisy $Y$.} to the ES-Cells approach\footnote{https://github.com/UW-AMO/TimeSeriesES-Cell}.
 HW  propagates outliers and is adversely affected by heteroscedastic noise, affecting the estimates of level, trend, and especially seasonality components (see Figure~\ref{fig:noisy} (b-d)). This lack of robustness gives a poor understanding of the overall time series (Figure~\ref{fig:noisy} (a)) and leads to low forecasting accuracy, as corrupted errors are propagated in future times (see Figure~\ref{fig:noisy} (e)). For RHW, automatic approaches to find $x_0, \alpha,\beta,\gamma$ failed, and we had to hand-tune parameters; the final result improves on HW but requires a long `burn-in' period, and still produces a somewhat noisy forecast.  The ES-Cells approach captures and removes outliers and heteroscedastic noise, and correctly identifies the components.

\section*{Time series estimation using ES-Cells}
The ES-Cells approach is constructed from interconnected building blocks. The basic cell consists of local ES estimation over a fixed window, equipped with a convex regularization term (for denoising) and a robust loss function (to guard against outliers), see Fig.~\ref{fig:approach} (a). %The time series components are assumed constant over the window.
The cells are then linked together by the time series dynamics, but allowing discrepancies between $x_t$ and $Ax_{t-1}$, see Fig.~\ref{fig:approach}(b). These differences are treated as samples of $g_t$, analyzed, and used to build forecasting confidence intervals. Fitting the entire ES-Cells model is a convex problem, and can be done at scale using  efficient methods for dynamic optimization~\cite{aravkin2016generalized,JMLR:v14:aravkin13a}.

\subsection{ES cell model}
First we formulate inference for a single ES cell. Given a time point $t$ and an integer $K$, we take a window of size $2K+1$ that includes all the points in the interval $[t-K, t+K]$. Some measurements can be missing; 
and no time point outside $[0,T]$ has measurements. To model these cases, we introduce indicator variables 
\[
d_t = \begin{cases}
0 & t \not\in[0,T] \; \mbox{or} \; y_t \mbox{ missing } \\
1 & t\in [0,T] \; \mbox{and } y_t \mbox{ observed. } 
\end{cases}
\]
We also define a unimodal sequence of weights $\alpha$, with 
$$0 < \alpha_{-K} < \dots < \alpha_{-1} < \alpha_0 > \alpha_1 > \dots \alpha_K > 0. $$
The estimate $\check x_t$ depends only on observations in the times $[\max(t-K,0), \min(t+K, T)]$, 
and is obtained by propagating the estimate
at the start of the window at time $t-K$ to the middle of the window  at time $t$, where $\alpha= \max_i\alpha_i = \alpha_0$:
\begin{equation}
\label{eq:hatcheck}
\check x_{t} = A^{K}\hat x_{t-K}.
\end{equation}
The estimate $\hat x_{t-K}$ solves the optimization problem  
\begin{equation}
\label{eq:singleCellObj}
 \min_x \sum_{r = -K}^K d_{t + r}\alpha_r |y_{r+t}-a_{r+K}^T x| + \lambda |b^Tx|, 
 \end{equation}
 where $a_{r+K} := A^{r+K-1}w$, and $b = [0, 0, 0, 1, -1, 0, \dots 0]^T$ extracts the difference of two seasonality components from the state $x$. 
 The objective function~\eqref{eq:singleCellObj} extends the classic ES approach~\eqref{eq:simpleObj} in three respects. 
\begin{enumerate}
\item The terms $d_t$ keep track of missing observations. 
\item The loss used to compare $y_{r+t}$ to $a_{r+K}^T x$ is robust to outliers. 
\item The term $|b^Tx|$ adds total variation regularization for the seasonality components. 
\end{enumerate}
The objective function~\eqref{eq:singleCellObj} is convex, as long as the loss and regularizer are both chosen to be convex. 
%With the current choice of $|\cdot|$,~\eqref{eq:singleCellObj} is a linear program. 

Before presenting the fully linked dynamic model, we rewrite~\eqref{eq:singleCellObj} more compactly, avoiding sums.
Define
\[
\begin{aligned}
 Y_{t-K} & = \begin{bmatrix} y_{t-K} & \dots & y_{t+K}\end{bmatrix}^T, 
\quad \sA = \begin{bmatrix} a_0 & \dots & a_{2K}\end{bmatrix} \\
 D_{t-K} & = \mbox{diag}\left( d_{t-K}\alpha_{-K}, \dots, d_{t}\alpha_{0}, \dots, d_{t+K}\alpha_K\right).
 \end{aligned}
 \]
We can now write~\eqref{eq:singleCellObj} as  
\begin{equation}
\label{eq:singleFull}
\hat x_{t-K} = \arg\min_x \|D_{t-K}(Y_{t-K} - \sA x)\|_1 + \lambda |b^Tx|.
\end{equation}

\subsection{Linking the ES-Cells}
Each cell estimate only depends on local data. 
To connect $\check x_t$ and $\check x_{t-1}$, we assume that the estimates satisfy  
\[
 \check x_{t} = A \check x_{t-1}  + g_t 
 \]
where, in contrast to the error term $g \epsilon_t$, $g_t$ are i.i.d. Gaussian errors. 
This is equivalent to adding the penalty 
\[
\|g_t\|_2^2 = \|A\check x_{t-1}  - \check x_t\|_2^2 = \|A^{K}(A\hat x_{t-(K+1)} - \hat x_{t-K})\|_2^2, 
\]
see~\eqref{eq:hatcheck}.
This links together objectives of form~\eqref{eq:singleFull} to generate a single objective 
over the entire sequence \\
${\bf x} = \{x_{-K}, \dots, x_{0},\dots,  x_{T-K}\}$:
\begin{equation}
\label{eq:Fullobj}
\begin{aligned}
\hat {\bf x} &= \arg\min_{{\bf x}} \sum_{t=-K}^{T-K}  \|D_{t}(Y_{t} - \sA x_{t})\|_1 + \lambda_1 |b^Tx_{t}| +\\
&  \qquad\qquad \sum_{t = -K}^{T-K-1}\lambda_2\|A^{K}(A x_{t} - x_{t+1})\|_2^2 
\end{aligned}
\end{equation}
The problem in~\eqref{eq:Fullobj} is nonsmooth but {\it convex}, and has dynamic structure. It has far more variables 
than the classic nonconvex ES formulation in~\eqref{eq:simpleObj}. Nonetheless, it can be efficiently solved at scale
using recent algorithms for generalized Kalman smoothing~\cite{aravkin2016generalized,JMLR:v14:aravkin13a}.
Given ${\bf \hat x}$, the final time series estimate ${\bf \check x}$ is given by  
\[ 
{\bf \check x} = \left\{ A^K \hat x_{-K}, \dots, A^K \hat x_0,  \dots A^K  \hat x_{T-K}\right\}.
\]
\begin{figure*}
\center
\hspace{-.5cm}
\begin{tabular}{cc}\\ 
\includegraphics[scale=0.45]{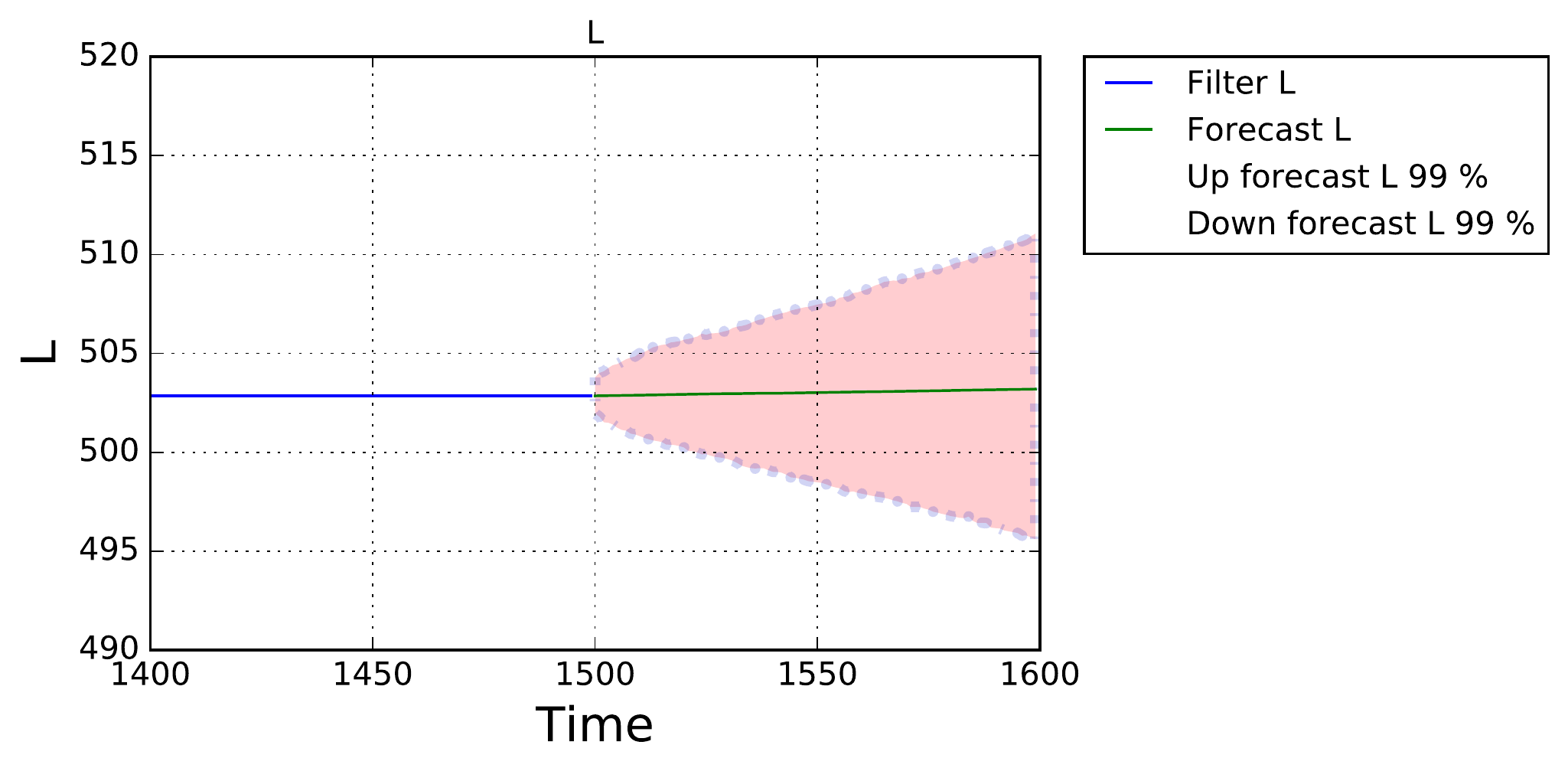}
& 
\includegraphics[scale=0.45]{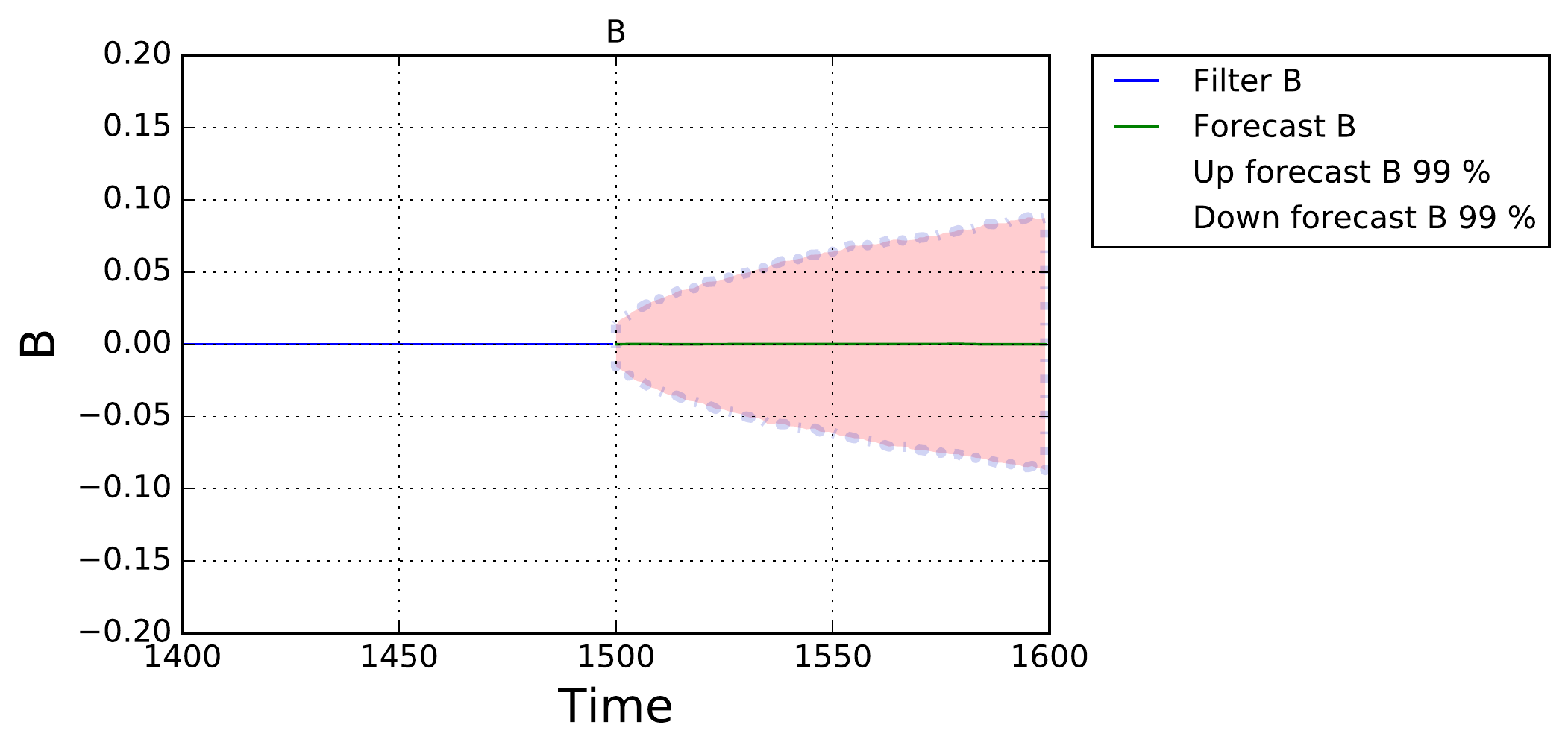}
\\ 
(a) {\bf Level: Forecast + 99\% CI. } 
\hspace{-.2in}
& 
\qquad (b) {\bf Trend: Forecast + 99\% CI. }\\
\qquad \includegraphics[scale=0.45]{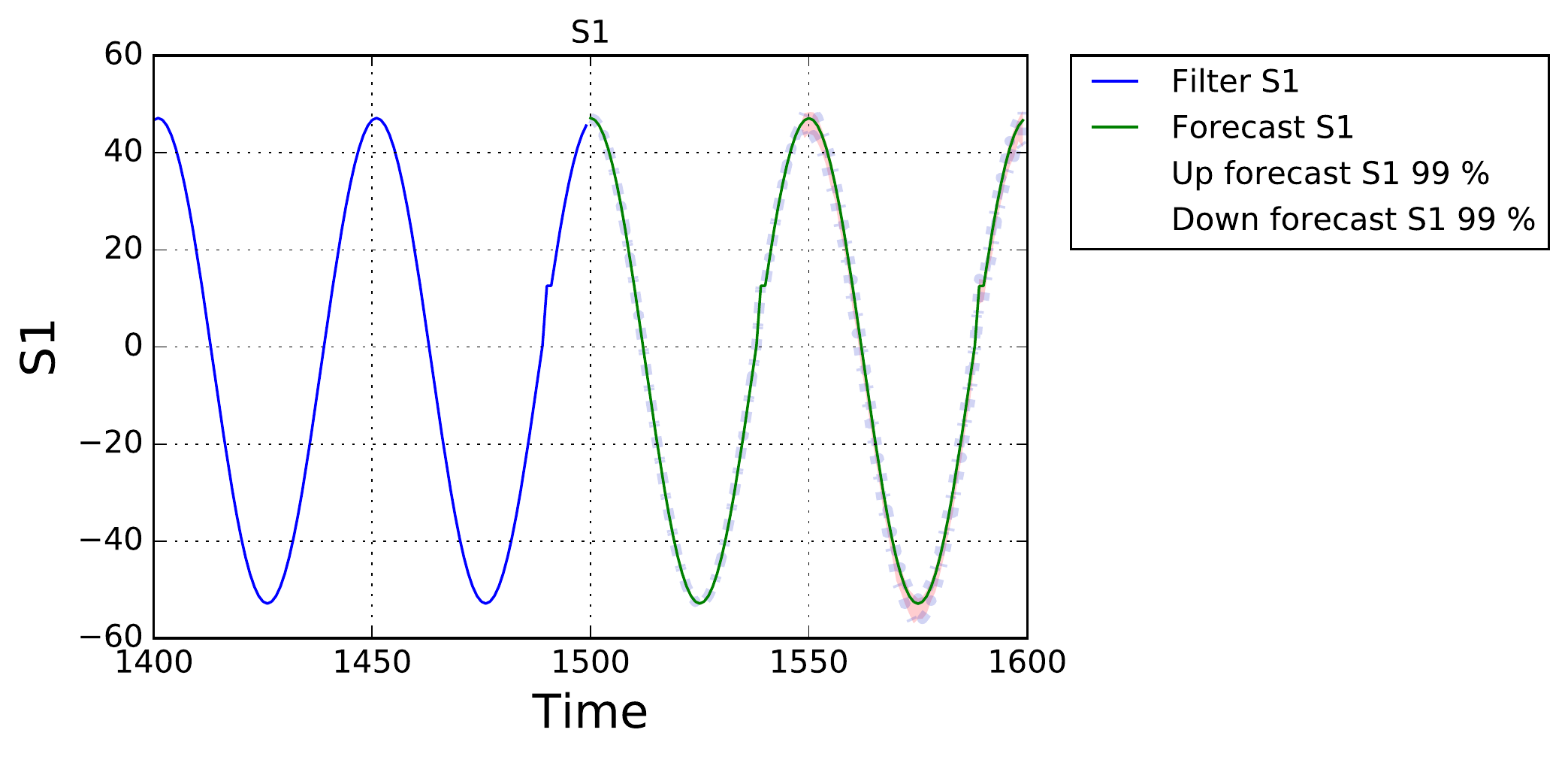}
&\qquad \includegraphics[scale=0.45]{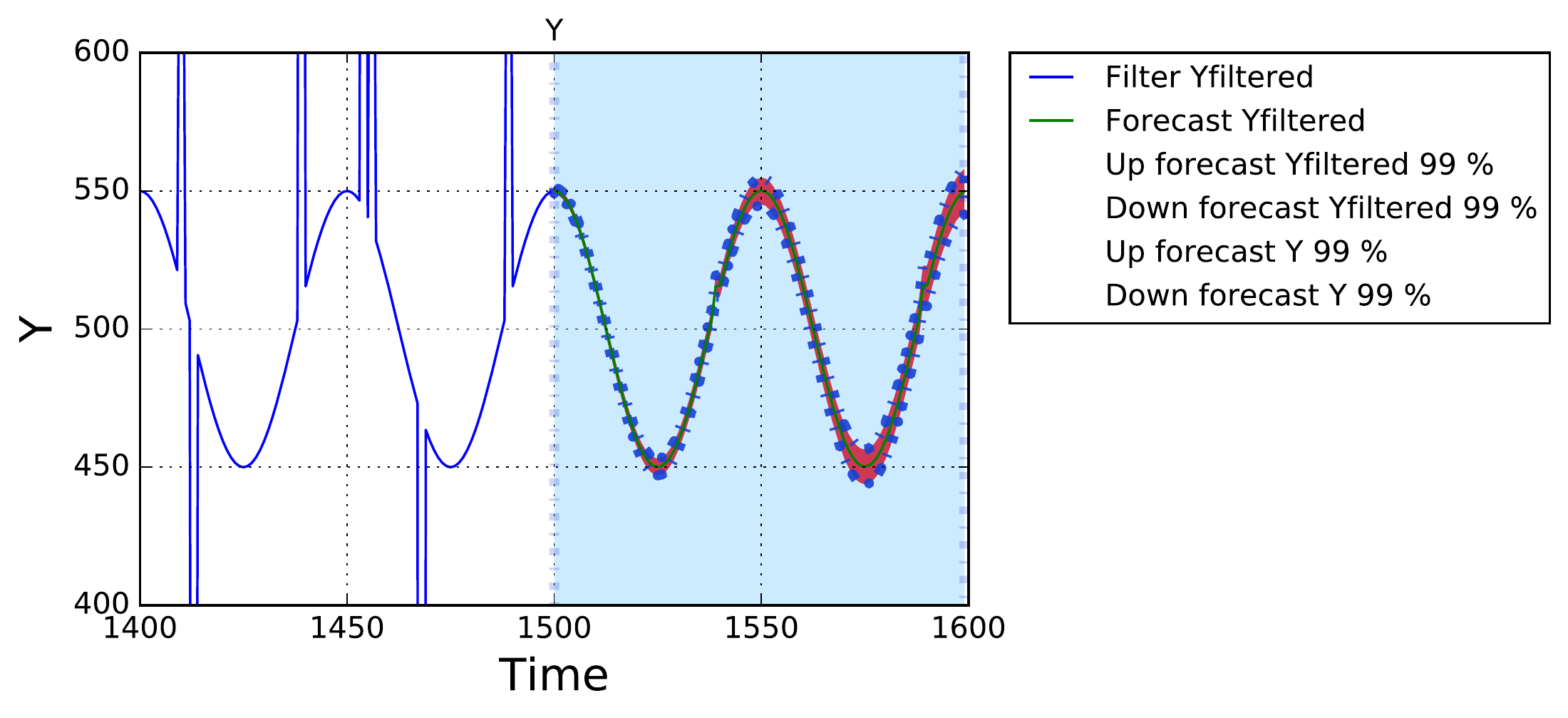}
\\
(c) {\bf Seas.: Forecast + 99\% CI. }  &\qquad (d) {\bf TS: Forecast + 99\% CI }.
%(e) {\bf Outliers detected by ES.}
\end{tabular}
\caption{\label{fig:forecast} Forecasting in the ES-Cells framework. Estimated smoothing parameters $g_t$ are sampled to 
provide 99\% confidence intervals for all components; panels (a), (b), (c) show forecasts for level, trend, and seasonality forecasts,
while (d) shows combined forecast for time series.  
Filtered signals are blue; mean forecast is green, with the 99\% CI shown in red. 
%Both $\epsilon_t$ and $g_t$ contribute 
%to the observation forecast (c) and (d). Inner CI is based on $g_t$ alone; outer CI is based on $\epsilon_t$, and so correctly 
%incorporates outlier occurrences.  
}  
%Right: overall approach, with multiple ES-Cells linked by a dynamic process model.} %(They are given in the text describing the experiment). }
\end{figure*}

\section*{Time series forecasting using ES-Cells}
ES-Cells capture two main sources of uncertainty that are important for forecasting future values of a time series: uncertainty in the residuals $\epsilon_t$, and in the smoothing parameters $\{ \alpha_t, \beta_t, \gamma_t \}$. 
ES-Cells track these two sources of uncertainty and can be used to create two separate confidence intervals: one representing the variability of each component of the signal, and the other capturing the structure of the residual.\\

Solving the full problem~\eqref{eq:Fullobj}, we obtain the entire sequence ${\bf \hat x}$, as well as corresponding estimates of residuals $\hat \epsilon_t$ and smoothing parameters $\hat g_t$: 
\[
\begin{aligned}
\hat \epsilon_t &= y_{t} - w^T\hat x_{t-1}\\
\hat g_t &= \hat x_t - A \hat x_{t-1}.
\end{aligned}
\]
In order to obtain the prediction distribution, we simulate sample paths from the models, using the 
empirical distribution of $\hat g_t$ and $\hat \epsilon_t$, and conditioned on the final state. 
This allows us to estimate any desired characteristics of the prediction distribution at a specific forecast horizon, and in particular to estimate confidence intervals that incorporate smoothing parameter and residual uncertainties. 
We can also incorporate model-based residuals (instead of using the empirical distribution) 
by generating forecasted $\epsilon_t$ values from any given distribution.

To illustrate the ES forecasting framework, Figure~\ref{fig:forecast} presents forecasts for the noisy synthetic model introduced in Figure~\ref{fig:noisy}. In particular, 100 step ahead forecasts and their 99\% confidence intervals (using 10000 Monte Carlo runs) are shown for trend and seasonality (panels (a) and (b)). These are obtained by using the empirical $\hat g_t$ distribution. The forecast for the full time series (and a zoomed plot) are shown in panels (c) and (d). The inner 99\% CI (strictly inside the shaded region) takes into account only uncertainty in smoothing parameters $g_t$,  while the outer CI (the border for the shaded region) takes into account uncertainty in $\epsilon_t$. Since the time series is contaminated by outliers, the outer CI is very wide in this case.

\section*{Real world Time Series : Twitter's user engagement dataset}
To test our algorithm we examine an anonymized time series representing user engagement on Twitter. This dataset is publicly available on its official blog\footnote{https://blog.twitter.com/2015/introducing-practical-and-robust-anomaly-detection-in-a-time-series} and is fully representative of the challenges tackled in this paper:

\begin{itemize}
\item Distinct seasonal patterns due to user behavior across different geographies
\item An underlying trend which could be interpreted as organic growth (new people engaging with the social network)
\item Anomalies or outliers due to either special events surrounding holidays (christmas, breaking news) or unwanted behavior (bots or spammers)
\item Underlying heteroscedastic noise. embodying the variance of the signal. 
\end{itemize}

\begin{figure*}
\center
\begin{tabular}{cc}\\ 
\includegraphics[scale=0.55]{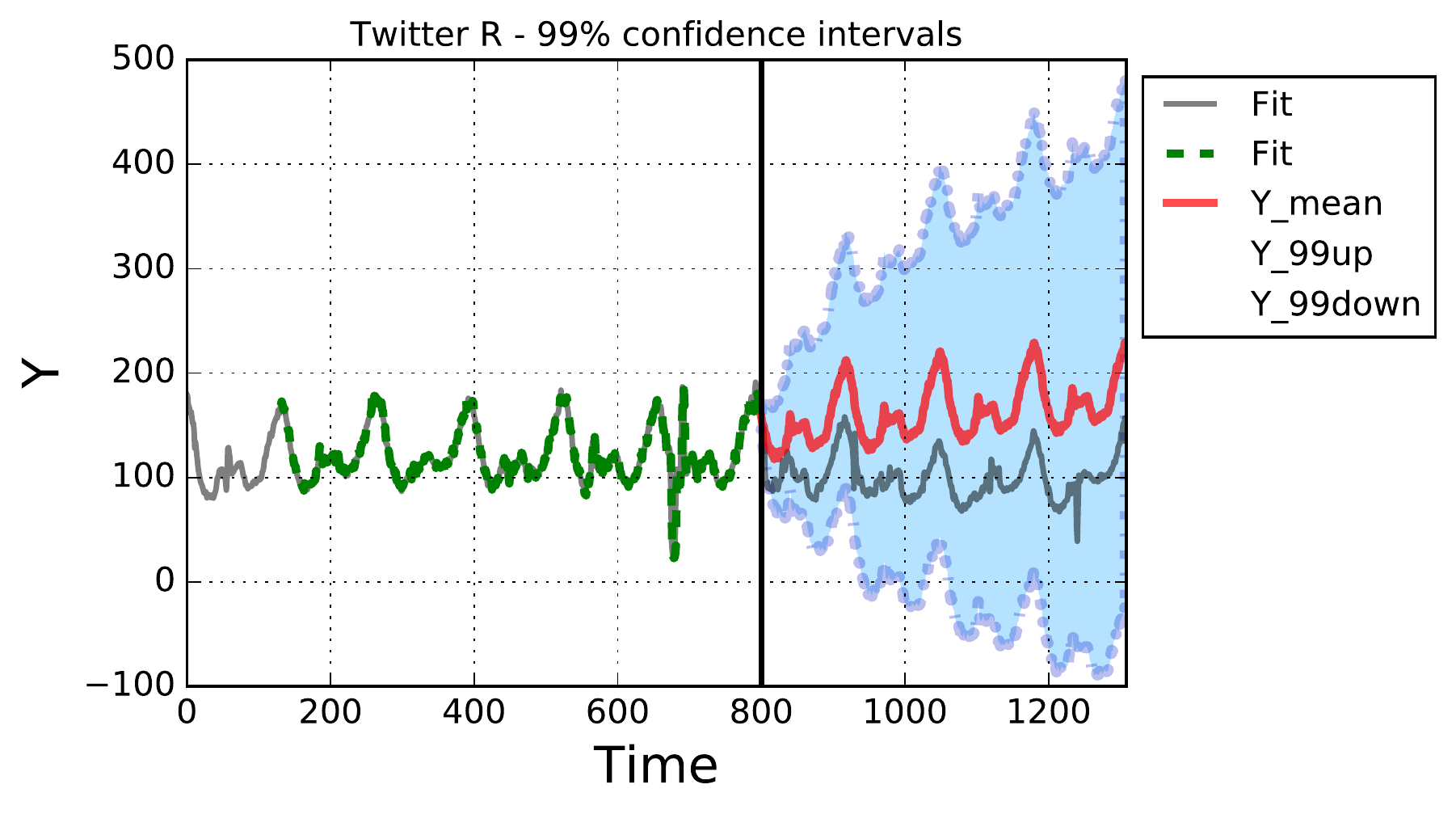}
&\includegraphics[scale=0.55]{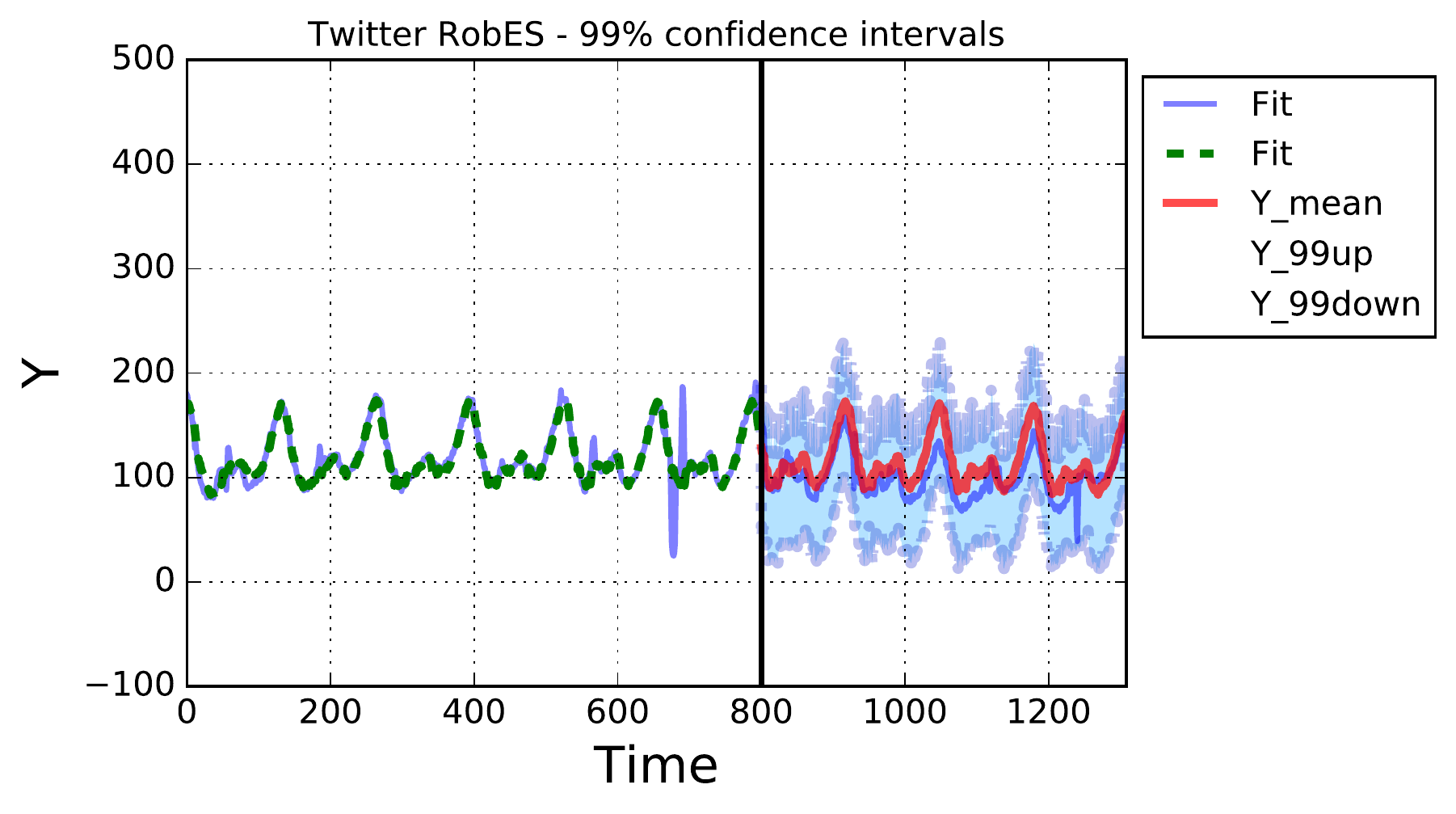}\\
(a) {\bf Classic Holt-Winters Analysis and Forecast} 
&(b) {\bf ES-Cells Analysis and Forecast.}
\end{tabular}
\caption{\label{fig:twitter} Analysis of Twitter user engagement data, using classic Holt-Winters Model (a) and ES-Cells approach (b). Classic approach (a) fits the outliers, obtaining very wide 99\% CI and, and forecasts sharp growth of the average user engagement. the ES-Cells approach does not fit the outliers; obtains tighter 99\% CI, and correctly forecasts a {\it decrease} in user engagement. Moreover, the 99\% CI in (b) is not symmetric; it is far tighter above than below.  } %(They are given in the text describing the experiment). }
\end{figure*}

\begin{figure}
\center
\includegraphics[scale=0.4]{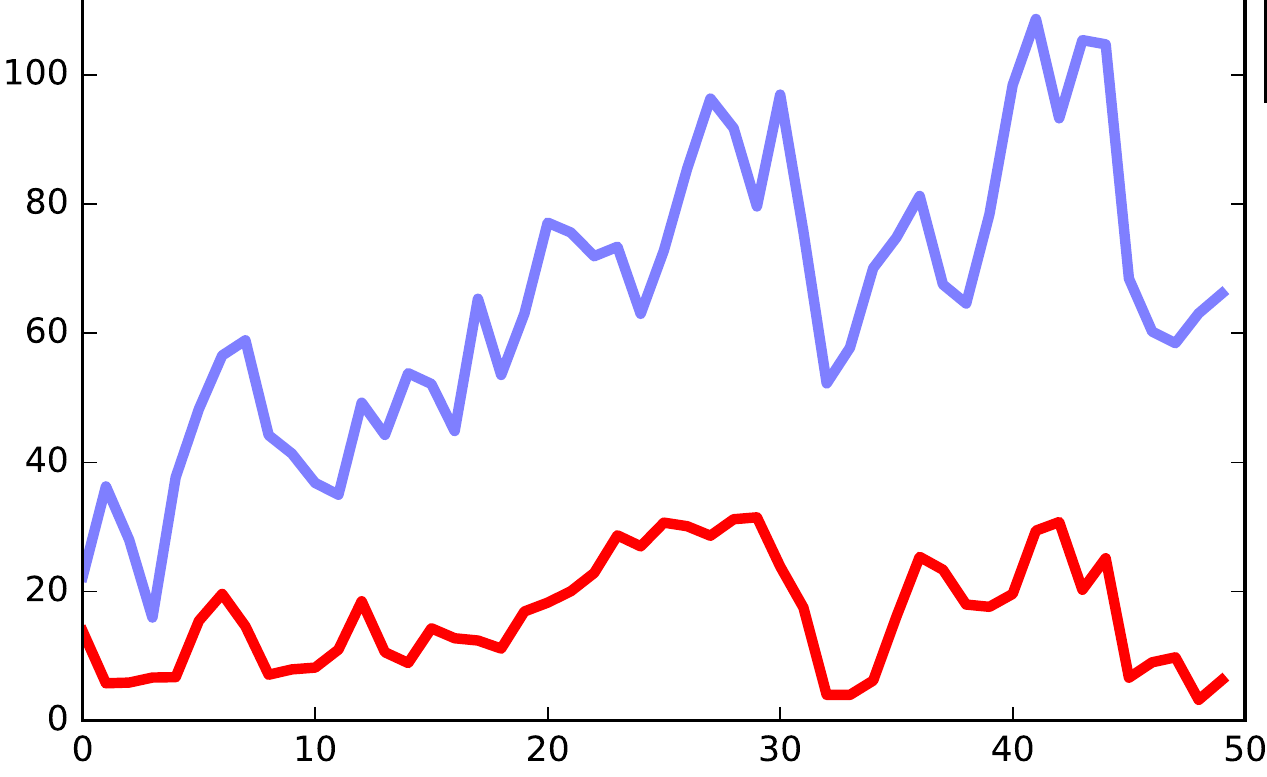}
\caption{\label{fig:mape} A comparison of the mean absolute percentage error (MAPE), to quantify the improvement of ES-Cells (red) over H-W (purple) for the twitter data in Figure~\ref{fig:twitter}.}
\end{figure}

The dataset was originally put online to showcase a robust anomaly detection procedure. With the ES-Cells framework, we can go much 
further, decomposing the time series into interpretable components, and then forecasting both the components and the entire time series under uncertainty. The original aim (anomaly detection) is easily accomplished by studying the tail of the empirical residual distribution, 
as discovered by the approach. 

%Using our algorithm we enlarge this paradigm by decomposing the time series into its principal components and forecasting them as well as the overall time series under uncertainty. We also show that our algorithm is able to single out outliers by looking at the tail of the residual distribution.

\begin{figure}
\center
\begin{tabular}{c}
\includegraphics[scale=0.4]{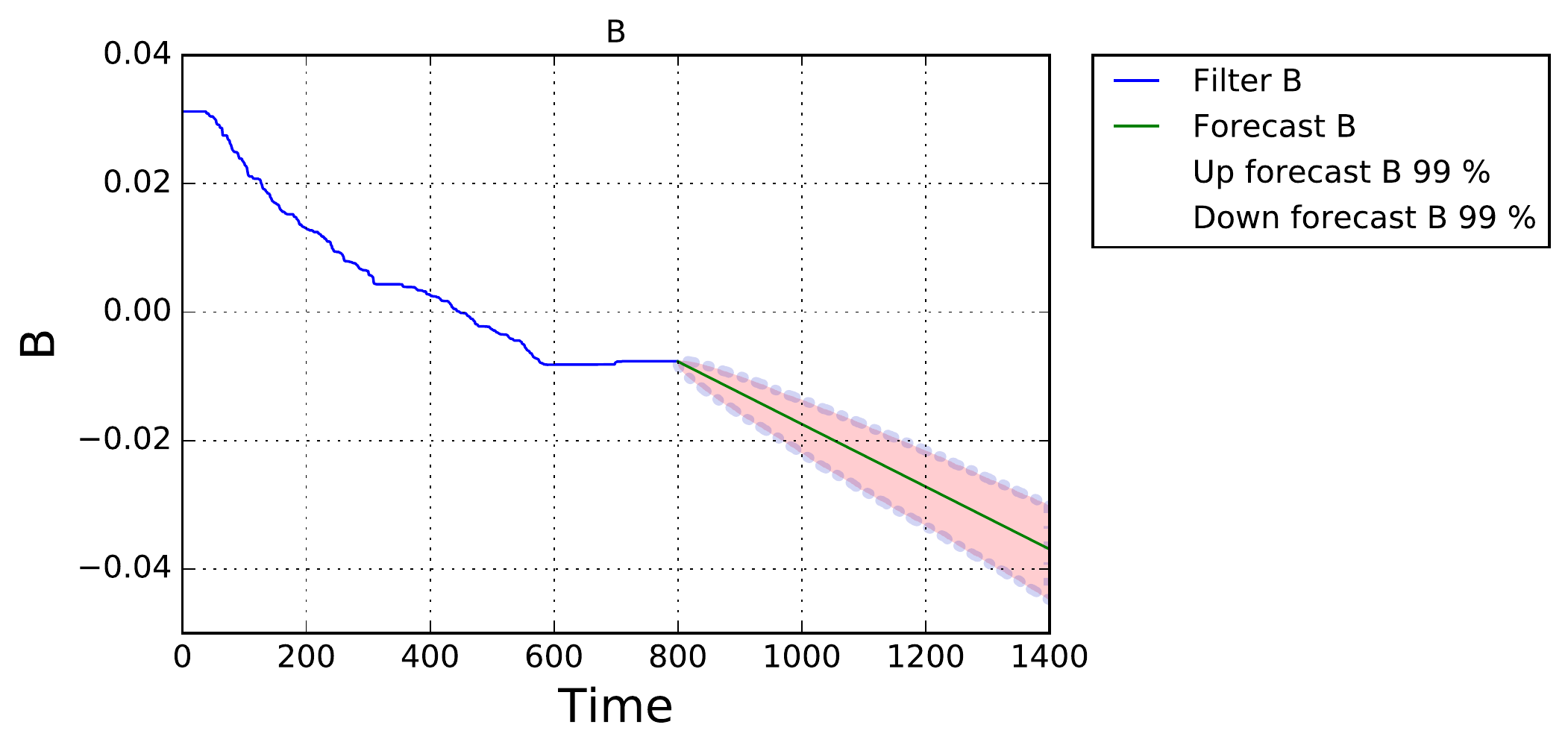}\\
(a) {\bf Trend, Forecast, \& 99\% CI. }\\ 
\includegraphics[scale=0.4]{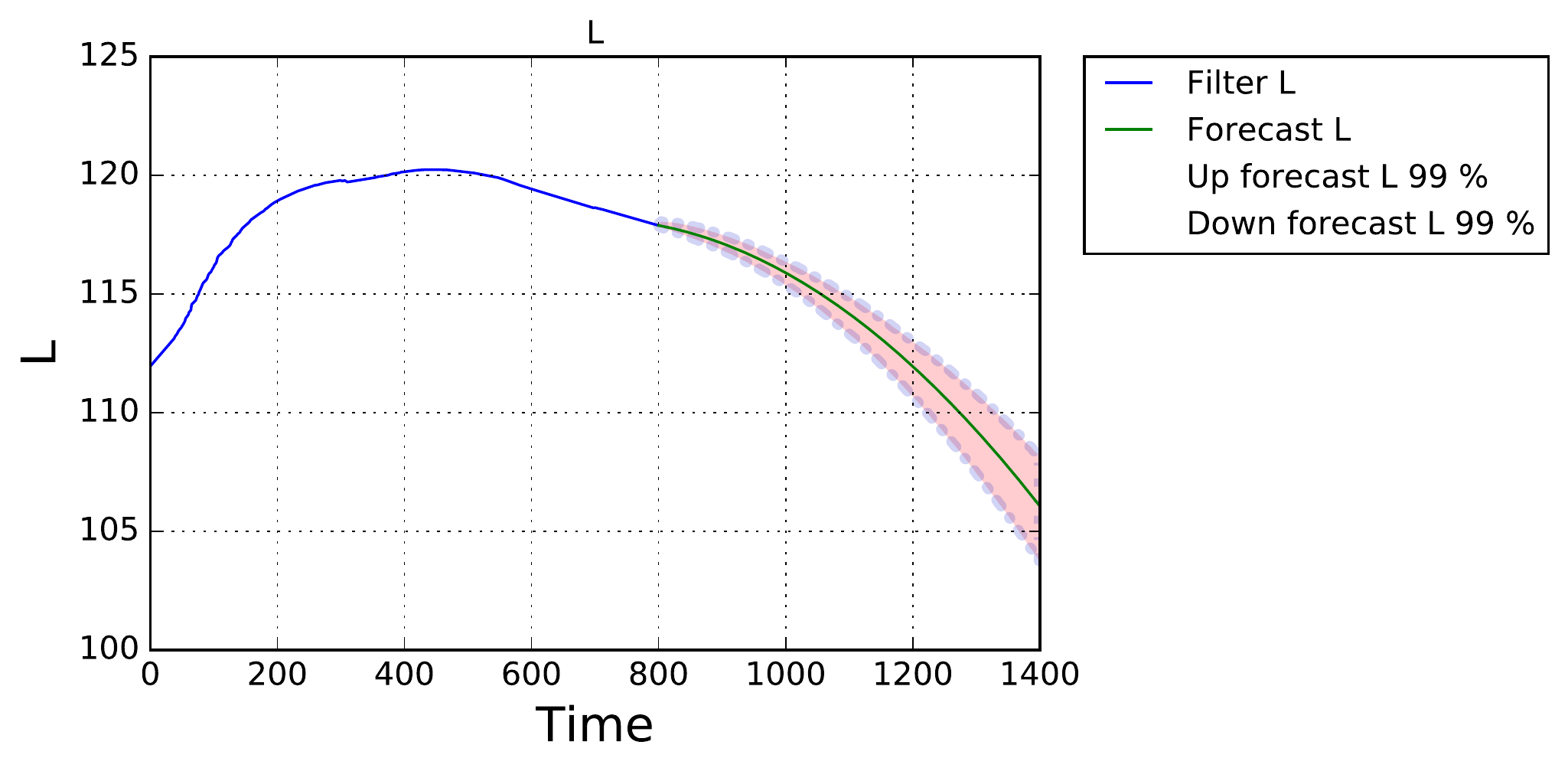}\\
 (b) {\bf Level, Forecast, \& 99\% CI. }\\
\includegraphics[scale=0.4]{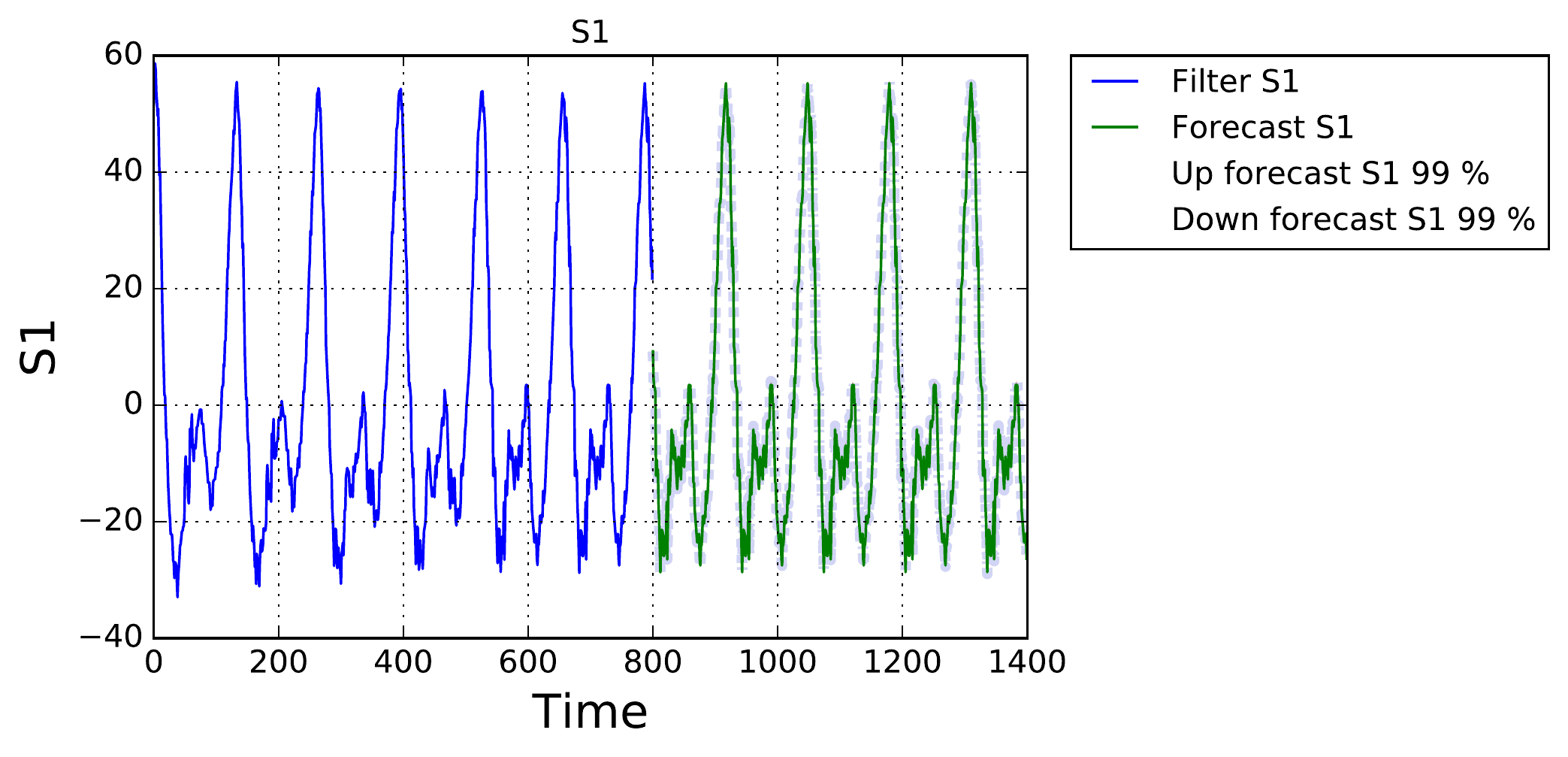}\\
(c) {\bf Seas., Forecast, \& 99\% CI}  %& (d) {\bf MAPE: H-W vs. ES-Cells}.
\end{tabular}
%\\
%(e) {\bf Outliers detected by ES.}
\caption{\label{fig:twitter_detail} Twitter dataset: Forecasts and 99\% CI for trend (a), level (b) and seasonality (c) obtained 
by the ES-Cells approach. There is a clear downward direction in level and trend.   }  
%Right: overall approach, with multiple ES-Cells linked by a dynamic process model.} %(They are given in the text describing the experiment). }
\end{figure}

%The main result is shown in Figure~\ref{fig:twitter}. From the results, it is easily seen that 
The classic Holt-Winters model fits outliers, forecasting sharp growth of engagement, which misses the observed trend (Figure~\ref{fig:twitter}(a)), and finds a very wide 99\% confidence interval, 
%that makes it seem like the series is too noisy to be forecasted with any accuracy. 
In contrast, the ES-Cells approach (Figure~\ref{fig:twitter}(b)) avoids fitting the outliers; the average forecast correctly captures the  {\it decrease} in the trend, and provides
a much tighter asymmetric 99\% CI. 
%This analysis gives a lot more information about the structure and behavior of the time series. 
The improvement can be quantitatively assessed by looking at the Mean Absolute Percentage Error (MAPE) for the forecasted time series over a sliding window of 10 observations, Figure~\ref{fig:mape}. Traditional Holt-Winters has a higher MAPE than ES-Cell at every time point; moreover, the MAPE of the ES-Cells method is stable over time, while the MAPE of ES increases, illustrating its failure to robustly capture the long term trend of the time series.
Trend, level and seasonality are shown in Figure~\ref{fig:twitter_detail}. There is a clear decrease in level and trend, which are detected despite the large amounts of noise in the data.

\begin{figure}
\center
\includegraphics[scale=0.5]{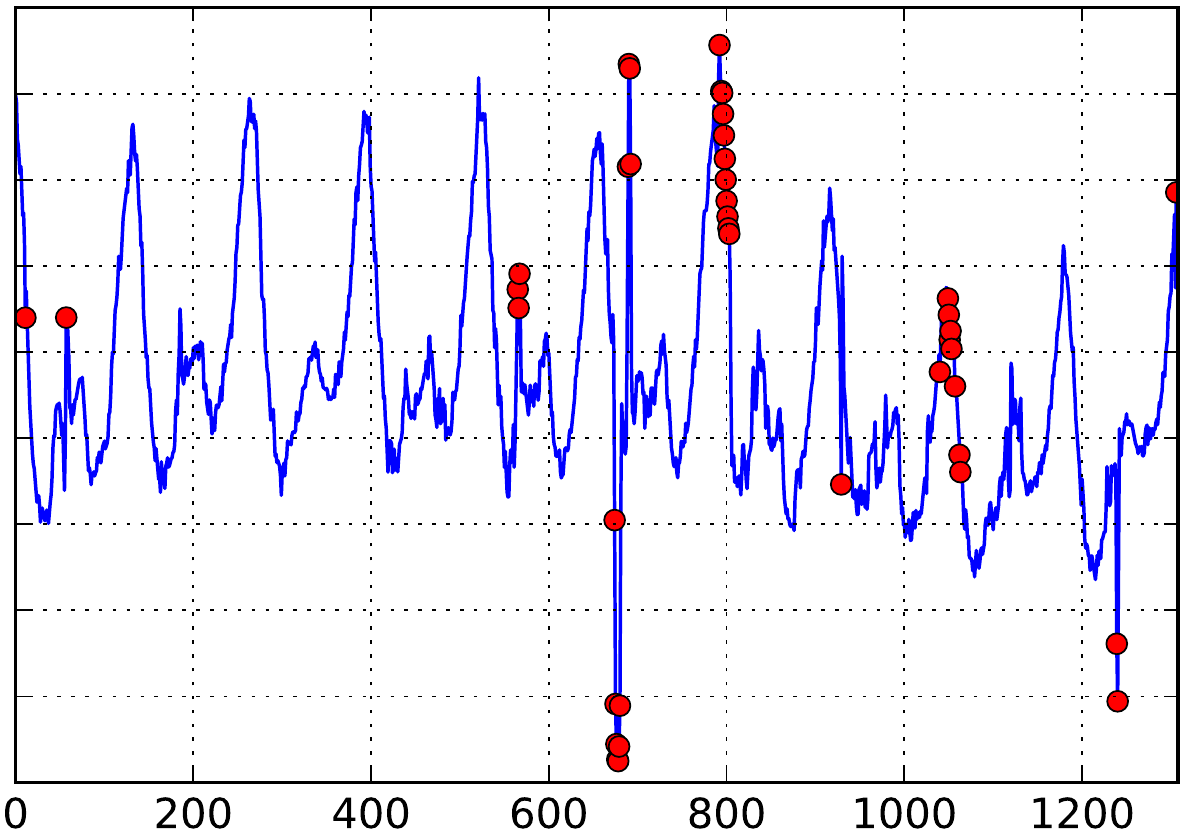}
\caption{\label{fig:outliers} Anomaly detection from the ES-Cells fit. The 1.5\% most extreme observations 
are highlighted using red dots.  }  
\end{figure}

%The seasonality 
%inference looks completely reasonable.
%the autocorrelation function before and after the ES analysis, 
%shows that the majority of the structure has been explained by the analysis~\ref{fig:twitter_auto}. 

\subsection{Anomaly Detection}

After fitting the ES procedure, we are left with a residual that we can analyze to understand anomalies in the time series. %Outliers and unusual behavior can be detected in the tails of the residual. 
Figure~\ref{fig:outliers} shows an example of outlier detection by looking at the 1.5\% tails of the residual distribution.

%\begin{figure}
%\center
%\includegraphics[scale=0.5]{plots/PlotPaper/camera_ready/AutocorrelationTwitterBeforeAfter}
%\caption{\label{fig:twitter_auto} Autocorrelation function of the signal, before (blue) and after (green) the ES-Cells decomposition. 
%It is clear that the majority of the coherent structure has been explained.  }  
%%Right: overall approach, with multiple ES-Cells linked by a dynamic process model.} %(They are given in the text describing the experiment). }
%\end{figure}

\vspace{0.2in}
\subsection{Robust auto completion of missing data}

\begin{figure}
\center
\includegraphics[scale=0.5]{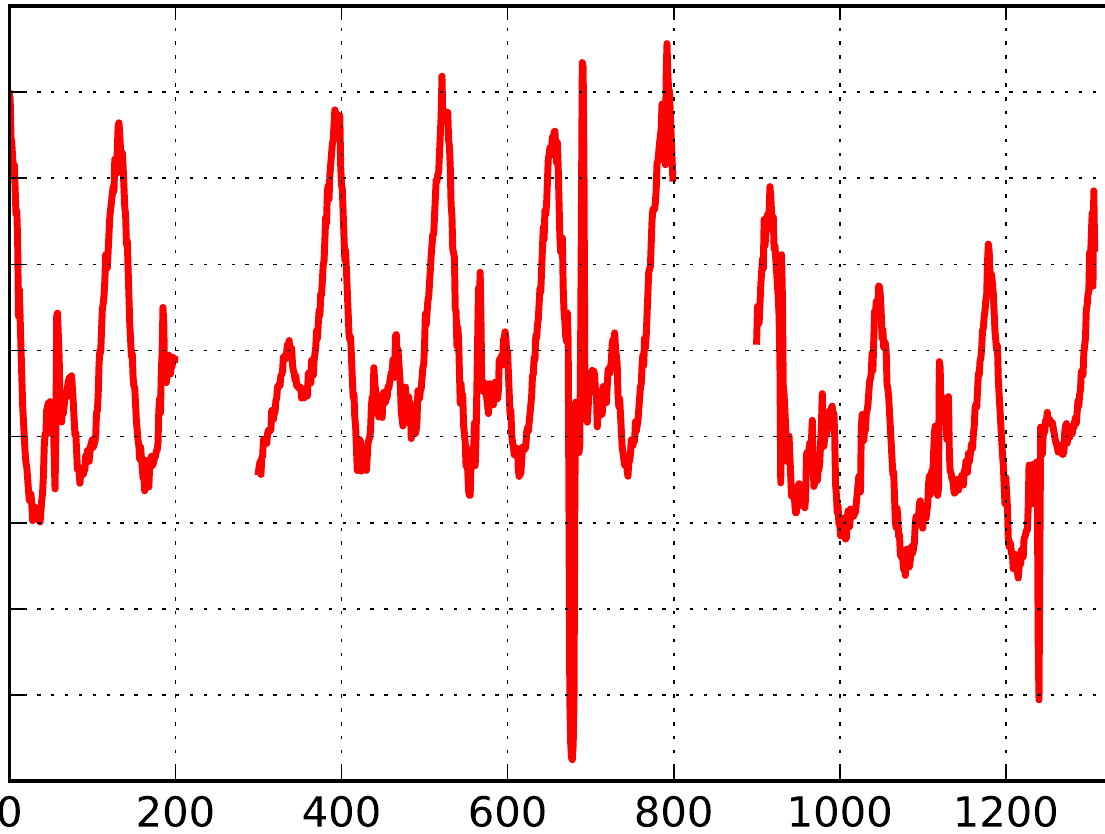}\\
(a) {\bf Deleted data} \\
\includegraphics[scale=0.5]{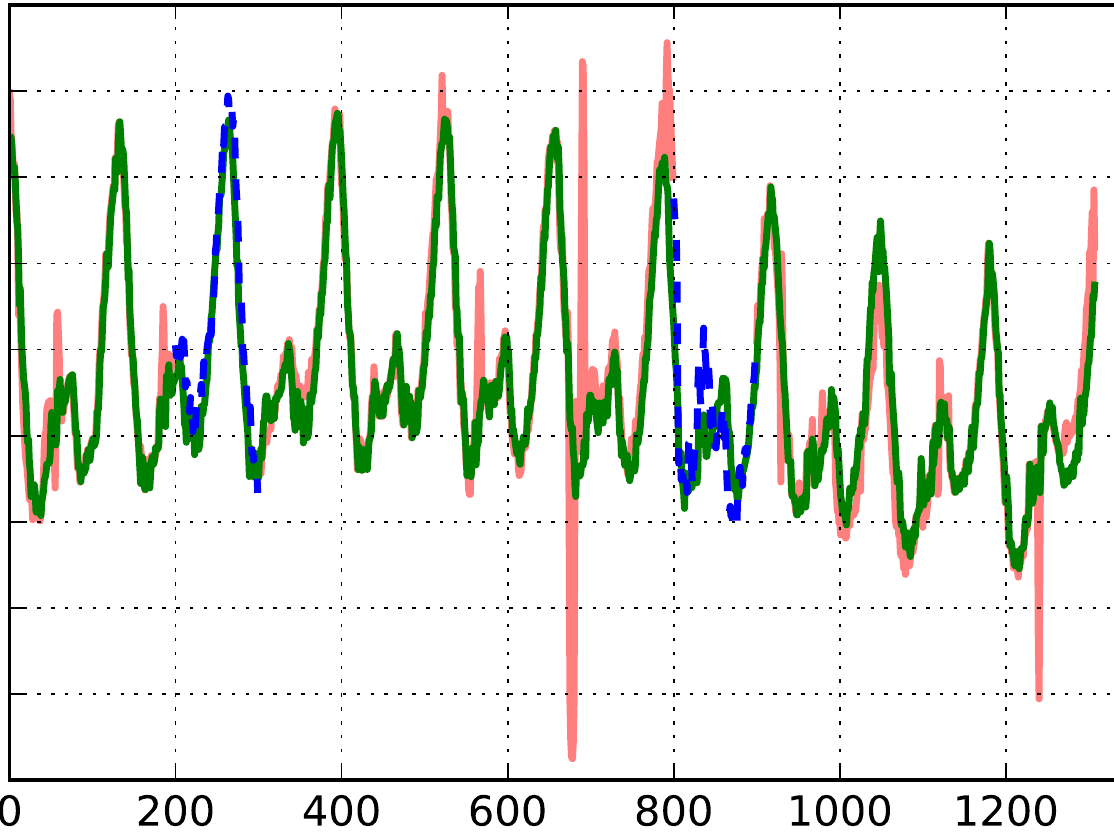}\\
(b) {\bf Auto fill-in.}
\caption{\label{fig:fill} ES-Cells can fill in data missing both at random and in contiguous chunks. In panel (a), two sections of 100 observations each have been deleted. Panel (b) shows the in-filled values (green) plotted over the data from panel (a) (red). The deleted data from panel (a) is shown in blue in panel (b).} %(They are given in the text describing the experiment). }
\end{figure}

The ES-Cell algorithm is also robust to missing observations. Whether the data is missing at random, or in significant contiguous batches, it is automatically filled in by the ES-Cells algorithm. Since the problem is solved globally, nearby outliers do not affect the interpolated values, in contrast to local interpolation methods. 
Figure~\ref{fig:fill} shows the result obtained by removing two contiguous chunks of 100 observations each in two distinct parts of the time series. The data is automatically `in-filled' using the ES-Cell procedure.

\begin{figure}
\center
\begin{tabular}{cc}\\ 
\hspace{-.1in}
\includegraphics[scale=0.23]{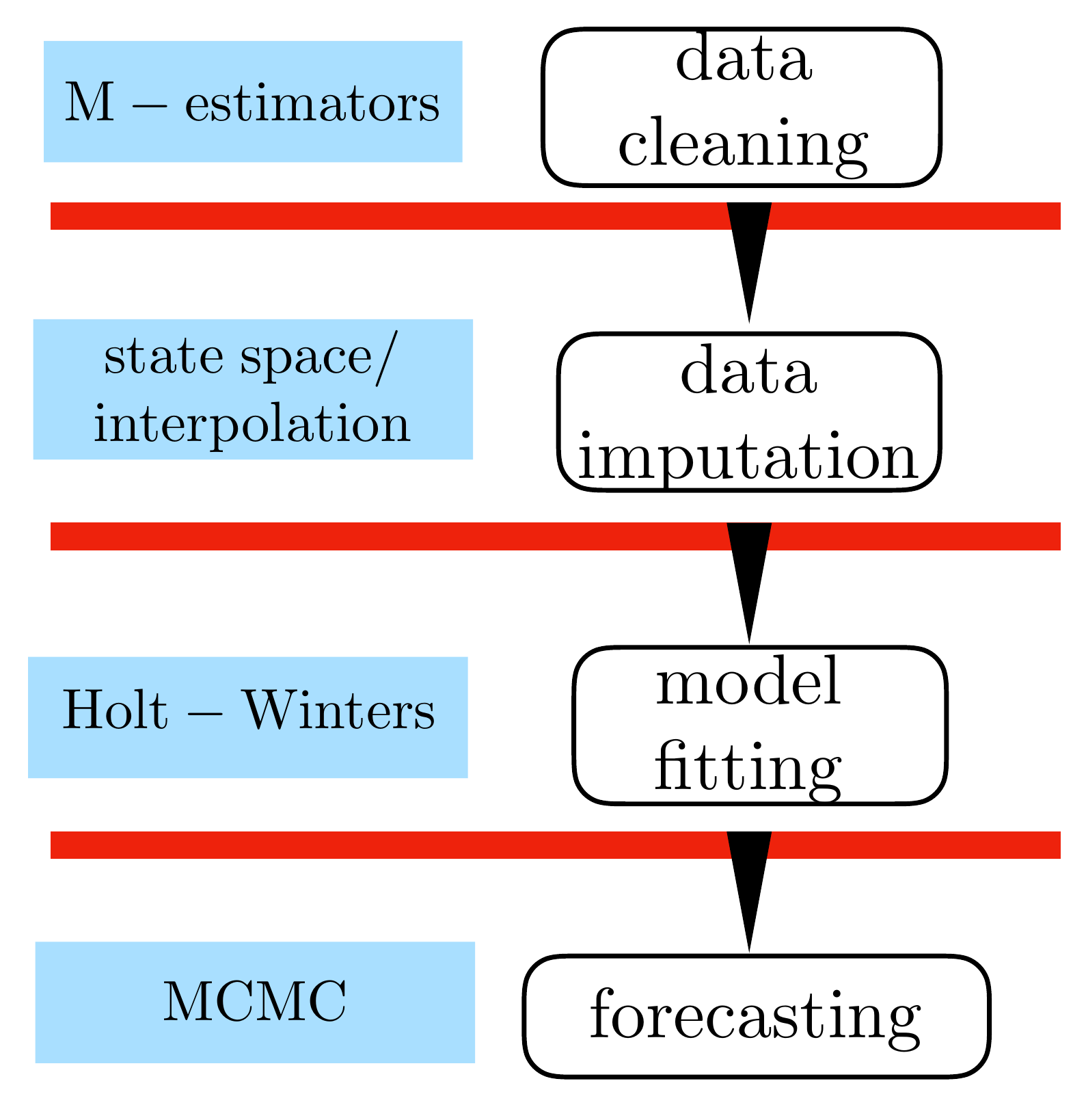}
\hspace{-.1in}
&\includegraphics[scale=0.23]{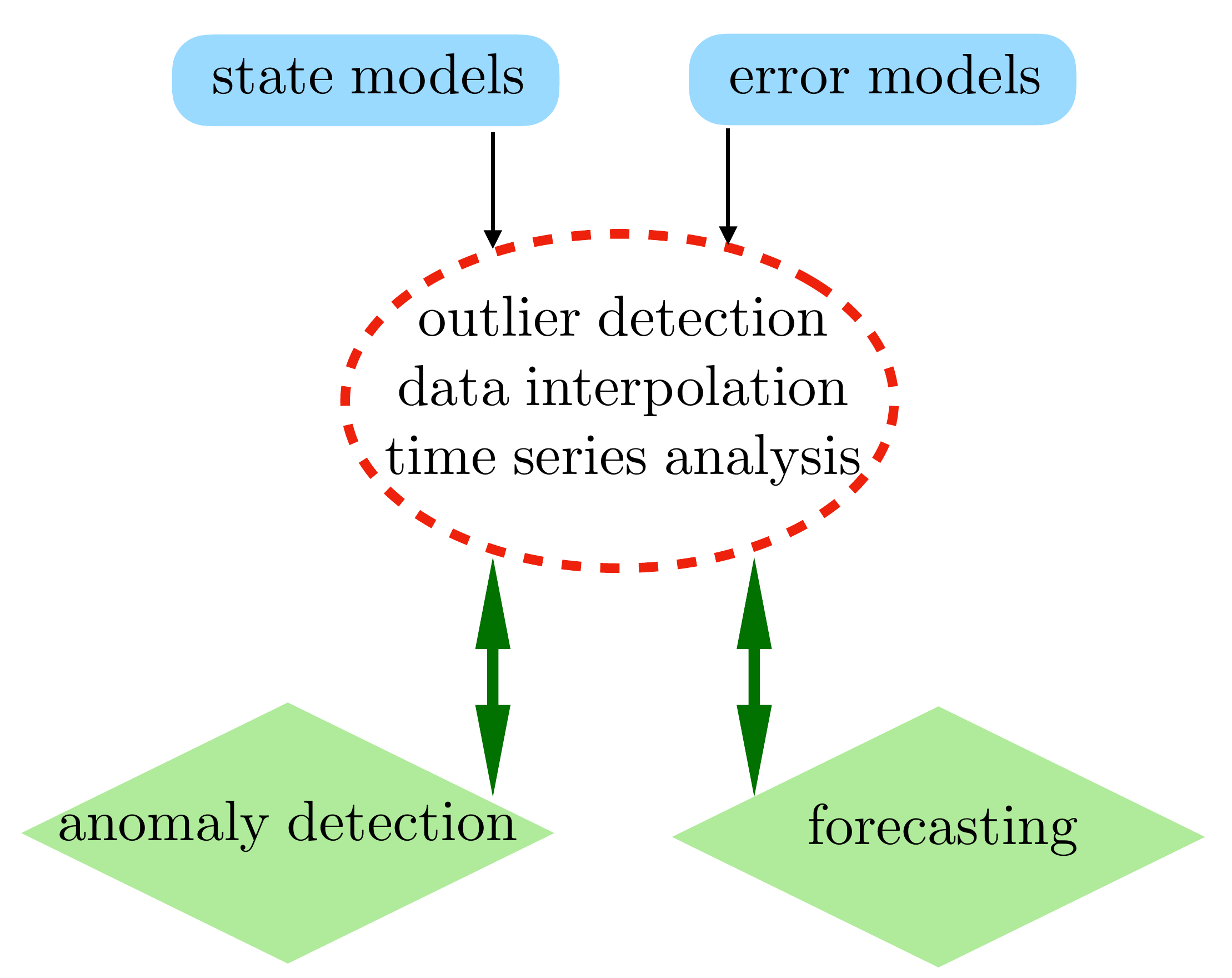}\\
(a) {\bf Classic approach} 
& (b) {\bf Global approach}
\end{tabular}
\caption{\label{fig:scheme} Classic TS analysis is sequential (a); data are pre-processed, then decomposed 
into components using models such as Holt-Winters. The approach is limited, because information about underlying structures such as level, trend, and seasonality is not available during the pre-processing. 
The new approach (b) is global; cleaning, interpolation, and decomposition are done in a unified context. 
Downstream applications, including anomaly detection and forecasting, significantly improve.} 
\end{figure}

\section*{Discussion}

ES-Cells is a new model for time series inference, capable of fitting and forecasting data
in situations with high noise, frequent outliers, 
and large contiguous portions of missing data. These features are present in many real-world large-scale datasets. 
The ES-Cells formulation differs from previous model in its global approach, as shown in Figure~\ref{fig:scheme}.
We simultaneously denoise, impute, and decompose the time series by solving a single 
convex optimization problem with dynamic structure; 
then use sampling-based strategies for forecasting and uncertainty quantification.  
The results are illustrated on simulated and real data, where the proposed method 
yields a 5-fold improvement in MAPE for the forecasting.  
The simplicity and versatility of the ES-Cells formulation makes it a superior 
alternative to Holt-Winters and related time series models. 
Code for the approach and experiments is publicly available\footnote{https://github.com/UW-AMO/TimeSeriesES-Cell}.

%\vspace{2in}

\subsubsection*{Acknowledgements}
The work of A. Aravkin was supported by the Washington Research Foundation Data Science Professorship.
\newpage
\clearpage
\bibliographystyle{abbrv}
\bibliography{references_sasha}

\end{document}